%% file: main.tex
\documentclass{article}

\usepackage{PRIMEarxiv}

\def\keywordname{{\bfseries \emph{Keywords}}}%
\def\keywords#1{\par\addvspace\medskipamount{\rightskip=0pt plus1cm
\def\and{\ifhmode\unskip\nobreak\fi\ $\cdot$
}\noindent\keywordname\enspace\ignorespaces#1\par}}

\usepackage[binary-units=true]{siunitx}  
\usepackage[utf8]{inputenc} 
\usepackage[T1]{fontenc}    
\usepackage[hidelinks]{hyperref}       
\usepackage{url}            
\usepackage{booktabs}       
\usepackage{amsfonts}       
\usepackage{nicefrac}       
\usepackage{microtype}      
\usepackage{lipsum}
\usepackage{fancyhdr}       
\usepackage{graphicx}       
\graphicspath{{media/}}     
\usepackage{multicol}
\usepackage[shortlabels]{enumitem}
\usepackage{subfig}
\usepackage{multirow}
\usepackage{listings}
\usepackage{xcolor}
\DeclareSIUnit\img{img}
\usepackage[multiple]{footmisc}

\newcommand{\pycode}[1]{{\small\texttt{#1}\normalsize}}

\title{Profiling and Improving the PyTorch Dataloader for high-latency Storage \\  \small{ * \\ A Technical Report}} 

\author{
  Ivan Svogor, Christian Eichenberger, Markus Spanring, Moritz Neun, Michael Kopp \\
  Institute of Advanced Research in Artificial Intelligence \\
  Untere Viaduktgasse 16, 1030 Vienna  \\
  Austria\\
  \texttt{\{name.lastname\}@iarai.ac.at} \\
}

\begin{document}
\maketitle

\begin{abstract}

A growing number of Machine Learning Frameworks recently made Deep Learning accessible to a wider audience of engineers, scientists, and practitioners, by allowing straightforward use of complex neural network architectures and algorithms. However, since deep learning is rapidly evolving, not only through theoretical advancements but also with respect to hardware and software engineering, ML frameworks often lose backward compatibility and introduce technical debt that can lead to bottlenecks and sub-optimal resource utilization. Moreover, the focus is in most cases not on deep learning engineering, but rather on new models and theoretical advancements. In this work, however, we focus on engineering, more specifically on the data loading pipeline in the PyTorch Framework. We designed a series of benchmarks that outline performance issues of certain steps in the data loading process. Our findings show that for classification tasks that involve loading many files, like images, the training wall-time can be significantly improved. With our new, modified \pycode{ConcurrentDataloader} we can reach improvements in GPU utilization and significantly reduce batch loading time, up to $12\times$. This allows for the use of the cloud-based, S3-like object storage for datasets, and have comparable training time as if datasets are stored on local drives. 

\end{abstract}


\input{chapters/ch1_intro}
\input{chapters/ch2_parallelism}

\input{chapters/ch3_dataset_dataloader}
\input{chapters/ch4_all_throughputs}

\input{chapters/ch5_disussion}
\input{chapters/ch6_conclusion}

\section*{Code Availability}
Our concurrent data loader is available under \url{https://github.com/iarai/concurrent-dataloader}.

\section*{Author Contributions Statement}
\addcontentsline{toc}{section}{Author Contributions Statement}

Roles according to CRediT (Contributor Roles Taxonomy)\footnote{\url{https://credit.niso.org/}}:
I.S. software, investigation, visualization, writing (original draft, review \& editing); Ch.E. software (early), writing (review \& editing); M.S. writing (review \& editing); M.N. validation, conceptualization; M.K. conceptualization.

\bibliographystyle{unsrt}  
\bibliography{main}  

\appendix
\input{chapters/appendix}

\end{document}

%% file: chapters/ch1_intro.tex
\section{Introduction}

Modern Machine Learning (ML) frameworks for deep learning (DL) come with a variety of out-of-the-box tools that help researchers and practitioners accelerate their work through a straightforward definition of neural network architectures, optimization methods, loss functions, logging, etc. 
By providing abstractions through multiple layers (e.g. storage, computing hardware, and even the neural network itself), ML frameworks allow end-users to focus on DL models, data reasoning and solving automation challenges.
That focus on rapid model development often puts engineering behind ML frameworks to a secondary place ~\cite{8990455}. For ML engineers, new frameworks raise concerns about potential bottlenecks, technical debt, and poor (but necessary) design choices, all of which can lead to inefficient training and solutions unfit for production environments. 
 
With the continuous demand for testing new ideas and the increasing size of datasets~\cite{Aizman2019HighPI}, the performance of training the models is lately capturing more attention due to limited and expensive resources.
Nowadays, for achieving competitive state-of-the-art performance in video, image and speech processing one needs to train models on petascale data~\cite{Aizman2019HighPI}, which does not only highlight the importance of the training performance, but also the necessity data loading performance.
Therefore, profiling deep learning code is necessary as it may uncover bottlenecks coming from an inefficient model implementation, the ML framework in use, or even the hardware. 

Broadly speaking, the training of any given DL model can be split into three main stages~\cite{8990455}, 1) communication, 2) data-loading and 3) computation.
Popular ML frameworks provide various profiling tools for inspecting the performance of a model for those stages.
That usually entails resource related parameters, e.g. GPU/CPU utilization, function-call execution time, or power consumption.
In this work, we use \textit{throughput} as a convenient and a straightforward metric to evaluate training efficiency, since the above mentioned parameters can often be overwhelming and convoluted to interpret.
With model throughput one gets an end-to-end performance of DL training, which can be defined in two different ways, a) processed data items per second, b) processed data size per second. 
Usually, when dealing with reading raw data items or copying data (to memory or GPU) a common unit is \si{\mega\bit\per\second}.
However, for the training phase throughput is oftentimes expressed in \si{\img\per\second}.
Since this work is intended to shine a light on both aspects, both units are used to make the results easy to understand and comparable.
With higher throughput, the training resource utilization should be increased, thus reducing the total training (wall) time.
That said, in this report we focus on the data loading efficiency using PyTorch~\cite{NEURIPS2019_9015}, and in particular on cloud based storage, containing clean ready-to-use datasets, accessible to anyone without the need (and possibility) to make local\footnote{In this report, we refer to local storage as local drives of the training machine or a network file system, not RAM or GPU memory} copies.

More specifically, our engineering research question in this report is the following:
\begin{quote}\it
Keeping storage format, preprocessing (augmentation) and batch size fixed, how can we achieve high throughput under different latency scenarios (fast local storage and higher-latency remote storage) by optimizing the Python data loading code?
\end{quote}

In the following sections, we address these questions by examining the data loading procedure in PyTorch.
Furthermore, we show results of a variety of benchmarks that ultimately result in a new, more efficient dataloader, compatible to the existing one. 

The contributions of this technical report are the following:
\begin{itemize}
    \item We show that data loading throughput can be increased by the introduction of within-batch parallelism in the PyTorch Dataloader on a vanilla vision dataset and model, in particular in high-latency settings as when data is loaded from remote storage.
    \item We provide two drop-in replacements for the PyTorch Dataloader \url{https://github.com/iarai/concurrent-dataloader}.
\end{itemize}

\subsection{Benchmarking setup and resources}

To address the aforementioned questions, a series of experiments to benchmark the throughput are presented, through which we highlight bottlenecks during training a DL model.
For that reason, it is important to have a systematic approach with a stable baseline.
In profiling ML models, there are no standard metrics to evaluate training performance. Instead, it is common to use a widely known ML model and dataset, which we also do in this work.
\subsubsection*{Machine learning framework}

We will be focusing on PyTorch~\cite{NEURIPS2019_9015}, and investigate and compare the performance to PyTorch Lightning~\cite{Falcon_PyTorch_Lightning_2019}, which is a lightweight PyTorch wrapper for high-performance AI research.
As a baseline for the conducted experiments we will use the ResNet-18~\cite{he2015deep} architecture for benchmarking purposes, and train it on the ImageNet ILSVRC 2012 dataset~\cite{deng2009imagenet}.
Furthermore, by \textit{default}, we refer to unmodified versions available on public repositories, for PyTorch\footnote{\url{https://github.com/pytorch/examples/blob/master/imagenet/main.py}}
 and PyTorch Lightning\footnote{\url{https://github.com/Lightning-AI/lightning/blob/1.5.2/pl_examples/domain_templates/imagenet.py}}.

\subsubsection*{Computing resources}

As we are dealing with performance benchmarking, it is important to consider the computing hardware. All the hardware used in this work is shown in \autoref{tab:compute}.
The majority of experiments are performed on the node in Datacenter 1.
However, to investigate the influence of networking and various file systems, we repeat some experiments in Datacenter 2, located in a different facility. 
Additionally, we also perform some experiments on Google Colab and AWS EC2 as a sanity-check (see Appendix, \autoref{apx:exhaustive-benchmark}).

\begin{table}[ht!]
\centering
\small
\begin{tabular}{@{}llllllll@{}}
\toprule
\multicolumn{4}{l}{\textbf{Datacenter 1}} &
  \multicolumn{4}{l}{\textbf{Datacenter 2}} \\ \midrule
\multicolumn{2}{l}{CPUs} &
  \multicolumn{2}{l}{2x Intel(R) Xeon(R) Gold 6146 CPU @ \SI{3.20}{\giga\hertz}} &
  \multicolumn{2}{l}{CPUs} &
  \multicolumn{2}{l}{2x Intel(R) Xeon(R) Gold 6244 CPU @ \SI{3.60}{\giga\hertz}}\\
\multicolumn{2}{l}{GPUs} &
  \multicolumn{2}{l}{8x Tesla V100-PCIE-32GB (only 1 was used)} &
  \multicolumn{2}{l}{GPUs} &
  \multicolumn{2}{l}{4x Tesla V100-SXM2-32GB (only 1 was used)} \\
\multicolumn{2}{l}{Storage} &
  \multicolumn{2}{l}{6x NVMe 7T, INTEL SSDPE2KE076T8} &
  \multicolumn{2}{l}{Storage} &
  \multicolumn{2}{l}{2x NVMe 11.7T, Micron\_9300\_MTFDHAL12T8TDR} \\
\multicolumn{2}{l}{RAM} &
  \multicolumn{2}{l}{754Gi (Samsung DDR4-2600)} &
  \multicolumn{2}{l}{RAM} &
  \multicolumn{2}{l}{754Gi (Hynix DDR4-2933)} \\ \midrule

\multicolumn{4}{l}{\textbf{Google Colab}} &
  \multicolumn{4}{l}{\textbf{AWS (p3.2xlarge) \footnote{\url{https://aws.amazon.com/ec2/instance-types/p3/}}}} \\ \midrule
\multicolumn{2}{l}{CPUs} &
  \multicolumn{2}{l}{2x Intel(R) Xeon(R) CPU @ \SI{2.30}{\giga\hertz}} &
  \multicolumn{2}{l}{vCPUs} &
  \multicolumn{2}{l}{8} \\
\multicolumn{2}{l}{GPUs} &
  \multicolumn{2}{l}{1x Tesla K80} &
  \multicolumn{2}{l}{GPUs} &
  \multicolumn{2}{l}{1x Tesla V100} \\
\multicolumn{2}{l}{RAM} &
  \multicolumn{2}{l}{13 GB} &
  \multicolumn{2}{l}{RAM} &
  \multicolumn{2}{l}{61 GB} \\
\multicolumn{4}{l}{} &
  \multicolumn{2}{l}{Storage} &
  \multicolumn{2}{l}{SSD} \\ \bottomrule
\end{tabular}
\vspace{2mm}
\caption{Computing platforms used for experiments}
\label{tab:compute}
\end{table}


\newpage

\subsection{Measurements and calculations}

Throughout this work we will be addressing three major metrics:
\begin{enumerate}[(a)]
    \item \textit{runtime}, which represents the time difference $(t_f - t_i)$, recorded as Unix timestamp, with $(t_i)$ marking the beginning of the experiment and $(t_f)$ the end.
    The beginning is considered as the time when the first batch is being loaded, and the ending is considered as the time when the training processs finished. 
    \item \textit{throughput} [\si{\img\per\second}], which represents the number of loaded data items (in this case images) during the training process.
    To put it more formally, consider a training dataset of $N$ items, $D=\{i_1, i_2, ..., i_N\}$ and the number of epochs $N_{epochs}$.
    The image throughput is calculated as $T_{imgs} = (N_{epochs} \cdot N)/(t_f - t_i)$.
    We may assume that epochs have a similar duration as measurements
    confirm.
    \item \textit{throughput } [\si{\mega\bit\per\second}], represents the size of loaded data items during the training process.
    The throughput in \si{\mega\bit\per\second} is obtained as $T_{mbits} = \left(\sum_{n=1}^{N} size(item_n) / (t_f - t_i) / 1024^2 \right) \cdot 8$, where $size$ is the function that returns the image size in bytes. 
\end{enumerate}

For the initial and final experiment, we also address the GPU memory and processing utilization, measured at \SI{10}{\hertz}, i.e. over a \SI{100}{\milli\second} timeframe\footnote{\url{https://developer.download.nvidia.com/compute/DCGM/docs/nvidia-smi-367.38.pdf}}, in a sidecar process. 

\subsection{Data loading pipeline}

\begin{figure}[!htb]
\includegraphics[width=0.85\textwidth]{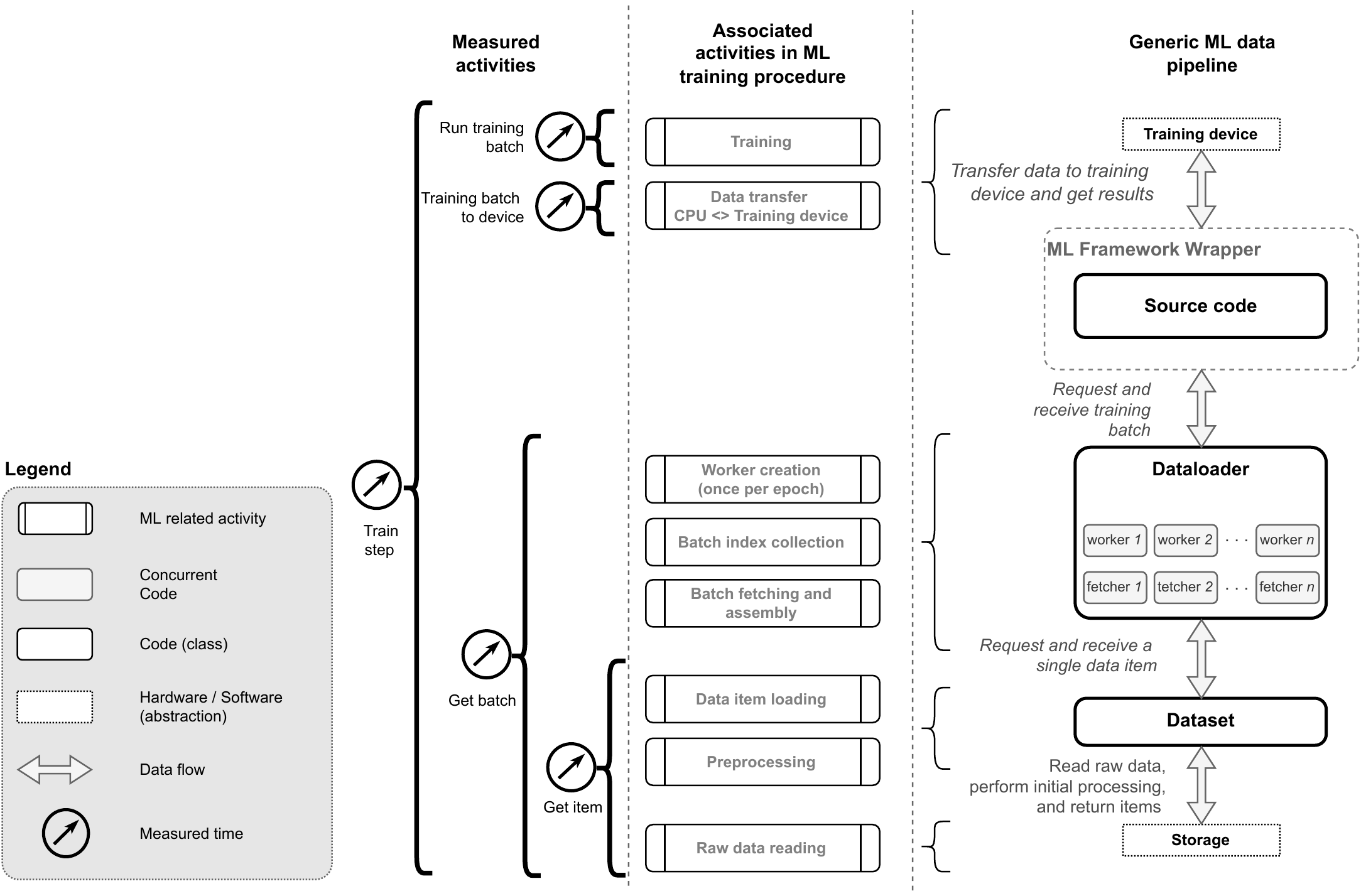}
\centering
\caption{Simplified overview of the data loading pipeline in Torch/Lightning.}
\label{fig:dataloader-stack-full}
\end{figure}

\autoref{fig:dataloader-stack-full} illustrates the typical data loading pipeline with PyTorch and PyTorch Lightning.
It highlights three important lanes, \textit{Measured activities}, \textit{Associated activities in ML training procedure}, \textit{Generic ML data pipeline} and.

From a software perspective, if we consider the layer that performs the training as the top layer, and the layer that collects data the bottom layer, then the Generic ML data pipeline with PyTorch usually consists of three main classes, a) the ML model itself, b) the \pycode{Dataloader} and c) the \pycode{Dataset}.
Looking at it from top-down, the ML model implements the model itself, its initialization, and the training loop, while it uses the \pycode{Dataloader} to trigger the batch loading process.
The middle lane of \autoref{fig:dataloader-stack-full} follows the software layers by describing associated activities in the ML training procedure.
We may notice that the Dataloader creates \pycode{Worker(s)}, that are class instances running as \pycode{Processes} in charge of creating batch index lists.
Each index of the batch list represents an index associated to the training item (in our case, an image) that will be fetched using a \pycode{Fetcher} class instance.
Furthermore, the \pycode{Dataset} implements the lowest level of the data loading process which fetches a single training item from the storage (regardless whether it is local or remote).
In our case, that means loading a single image, and returning it to the \pycode{Fetch(er)}. From there, it is forwarded to the \pycode{Worker}, which then assembles a collected batch and returns it to the training loop.
After the batch has been assembled, the data gets transferred to the training device and the training process can start. This process continues until all the items in the dataset are exhausted, and repeated for each epoch.

The left most lane, shows measurement points, i.e. parts of the code which we associate with specific and unique log entries so that we can extract the runtime, and visualize the execution order of main functions during the training phase.
Therefore, the \textit{Get batch} is associated with a \pycode{batch} log entry, and the function in charge for starting the batch loading process, \pycode{next\_data} is responsible for triggering the data loading process.
Considering the logged time, this includes the entire batch loading time, and the time to initialize this process (creating workers, fetchers, loading batch indexes, etc.), and loading a single item.
Furthermore, \textit{Get item} is associated with the log entry and function \pycode{\_\_getitem\_\_} implemented in the \pycode{Dataset} itself, and the logged time for it includes the time necessary to load the image from storage. If it is local, then one simply reads an image, however, if it is remote this time also include network-related overheads.
Once the batch is returned to the main ML model class, using the log entry \pycode{training\_batch\_to\_device} time to transfer the data to the GPU is logged, and finally, the time to run a training batch is logged. 

\subsection{Motivational experiment}

In this section we introduce a motivational experiment to empirically verify whether and where we have bottlenecks in training using PyTorch. Let's consider the aforementioned Vanilla implementation of ResNet-18 in PyTorch and PyTorch Lightning, with the following parameters:

\begin{table}[htb!]
\centering
\resizebox{0.48\textwidth}{!}{
\begin{tabular}{@{}lllllllll@{}}
\toprule
\textbf{\begin{tabular}[c]{@{}l@{}}Batch \\ size\end{tabular}} &
  \textbf{Workers} &
  \textbf{\begin{tabular}[c]{@{}l@{}}Dataset\\ limit (size)\end{tabular}} &
  \textbf{\begin{tabular}[c]{@{}l@{}}Learning\\ rate\end{tabular}} &
  \textbf{\begin{tabular}[c]{@{}l@{}}Weight\\ decay\end{tabular}} &
  \textbf{Epochs} \\ \midrule
\multicolumn{1}{c}{256} &
  \multicolumn{1}{c}{4} &
  \multicolumn{1}{c}{15000} &
  \multicolumn{1}{c}{0.1} &
  \multicolumn{1}{c}{0.0001} &
  \multicolumn{1}{c}{5} \\ \bottomrule
\end{tabular}
}
\caption{Parameters for the motivational experiment}
\label{tab:motivational-experiment-params}
\end{table}

As mentioned, we use the ImageNet dataset, which consists of 14 million images, with average dimension of 469x387\footnote{\url{https://towardsdatascience.com/compression-in-the-imagenet-dataset-34c56d14d463}}. However, for the experiments presented in this work, we use a reduced set of images (\pycode{dataset\_limit} parameter), and in the \pycode{Dataset} the following data augmentation (i.e. transform): 1) random resized crop to the dimension of 224x224, 2) perform a horizontal flip, 3) convert to tensor, and 4) normalize. 

The experiment was performed on the Datacenter 2 machine, using a single Nvidia Tesla V100 (SXM2-32GB) GPU as the training device, and we used \textit{scratch} storage (locally mounted SSDs). The results of measurements are shown in \autoref{tab:intro_motivation_example}.

\begin{table}[htb!]
\centering
\resizebox{\textwidth}{!}{
\small
\begin{tabular}{@{}ccrrrrrrr@{}}
\toprule
\textbf{Storage} & \textbf{Lib.} & \textbf{\begin{tabular}[c]{@{}c@{}}$GPU_{util=0}$ \\[5pt] {[\si{\percent}]}\end{tabular}} & \textbf{\begin{tabular}[c]{@{}c@{}}$GPU_{util>0}$\\[5pt] {[\si{\percent}]}\end{tabular}} &
\textbf{\begin{tabular}[c]{@{}c@{}}$GPU_{util=0}^{mem}$\\[5pt]  {[\si{\percent}]}\end{tabular}} &
\textbf{\begin{tabular}[c]{@{}c@{}}$GPU_{util>0}^{mem}$\\[5pt] {[\si{\percent}]}\end{tabular}} &
\textbf{\begin{tabular}[c]{@{}c@{}}Runtime\\[5pt] {[\si{\second}]}\end{tabular}} &
\textbf{\begin{tabular}[c]{@{}c@{}}Throughput\\[5pt] {[\si{\img\per\second}]}\end{tabular}} &
\textbf{\begin{tabular}[c]{@{}c@{}}Throughput \\[5pt] {[\si{\mega\bit\per\second}]}\end{tabular}} \\ \midrule
scratch & Torch & 26.08 & 74.78 & 29.67 & 41.83 & 137.24 & 546.48 & 492.95 \\
scratch & Lightning & 66.41 & 64.68 & 5.79 & 18.93 & 491.06 & 152.73 & 137.77 \\
s3 & Torch     & 95.44 & 71.75 & 1.77 & 41.06 & 2309.99 & 32.47 & 29.31 \\
s3 & Lightning & 98.04 & 65.28 & 0.34 & 19.19 & 8934.94 & 8.39  & 7.58  \\ \bottomrule
\end{tabular}
} 
\vspace{2mm}
\caption{Initial benchmark results, a comparison between PyTorch and Lightning using local storage (scratch), and remote storage (s3). Parameters according to Table~\ref{tab:motivational-experiment-params}.}
\label{tab:intro_motivation_example}
\end{table}
\vspace{15mm}
In addition to the previously explained runtime and throughput, the table also shows four columns associated with the GPU:
\begin{itemize}
    \setlength\itemsep{0em}
    \item $GPU_{util=0}$, the percentage of experiment runtime where the GPU is not used. 
    \item $GPU_{util>0}$, average GPU utilization (when GPU is not idle).
    \item $GPU_{util=0}^{mem}$,  the percentage of experiment runtime where the GPU memory is not used. 
    \item $GPU_{util>0}^{mem}$, average GPU memory utilization (when GPU is not idle).
\end{itemize}

For this experiment the GPU utilization is reported at a rate of \SI{10}{\hertz}, i.e. by averaging the utilization over a period of \SI{100}{\milli\second}.
The experiment measuring the runtime and GPU utilization for the Torch implementation reported an average utilization of \SI{74.78}{\percent}.
However, \SI{26.08}{\percent} of the overall experiment runtime the GPU was idle.
Furthermore, GPU memory is not used  \SI{29.67}{\percent} of the experiment and is utilized on average with \SI{41.83}{\percent}.
For Lightning, the GPU idle time seems much larger, as well as GPU memory utilization lower.
This indicates that there are significant overheads compared to the pure PyTorch implementation and there is room for improvement and fine-tuning since Lightning uses PyTorch "under the hood". 

This experiment shows that there is considerable room for improvement if one is able to reduce the time in which the GPU is idle, namely \SI{26}{\percent} for PyTorch and \SI{66}{\percent} for Lightning.

\begin{figure}[htb!]
	\begin{minipage}{0.40\linewidth}
        \centering
        \includegraphics[width=\textwidth, trim={0 1cm 0 0}, clip]{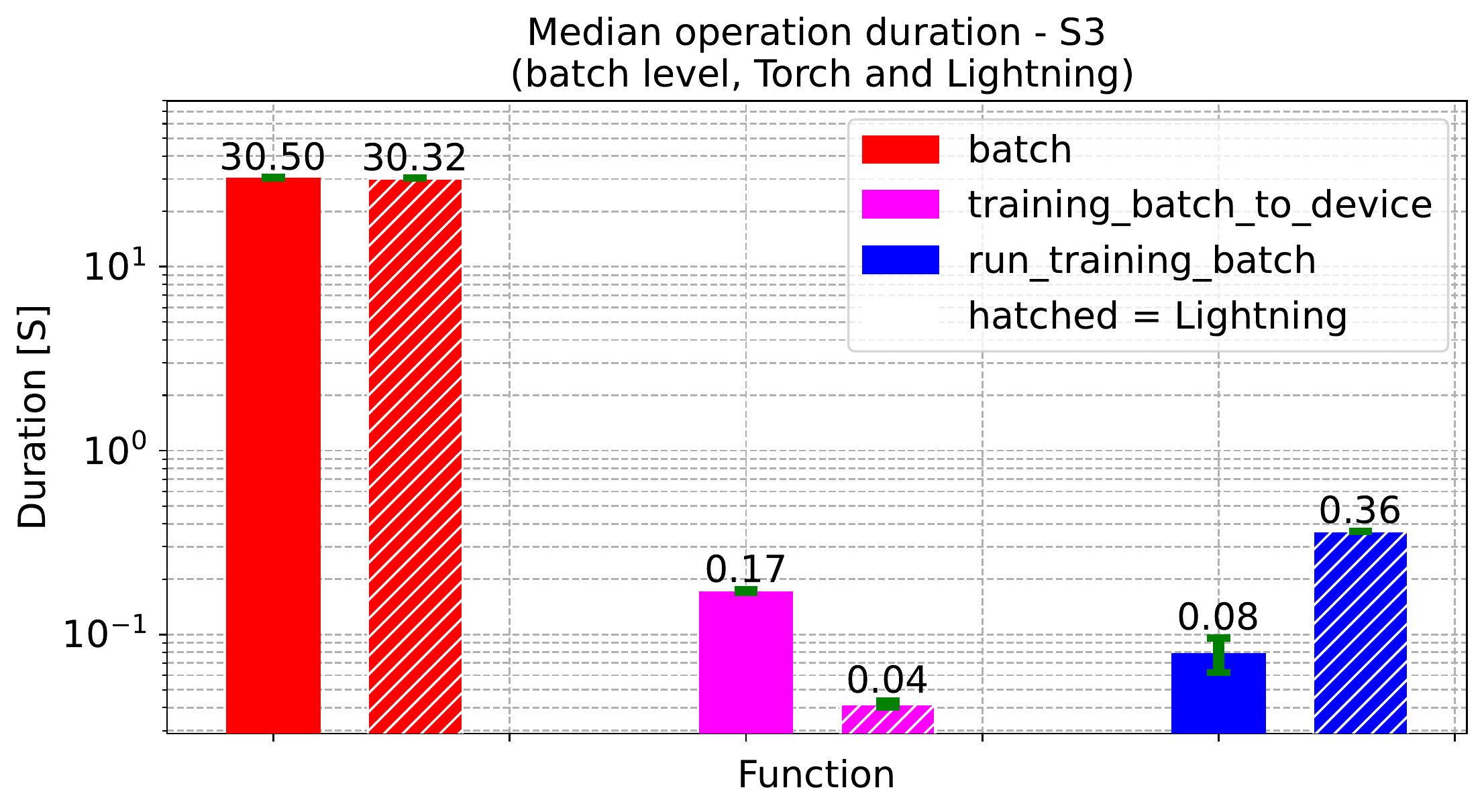} 
        \includegraphics[width=\textwidth]{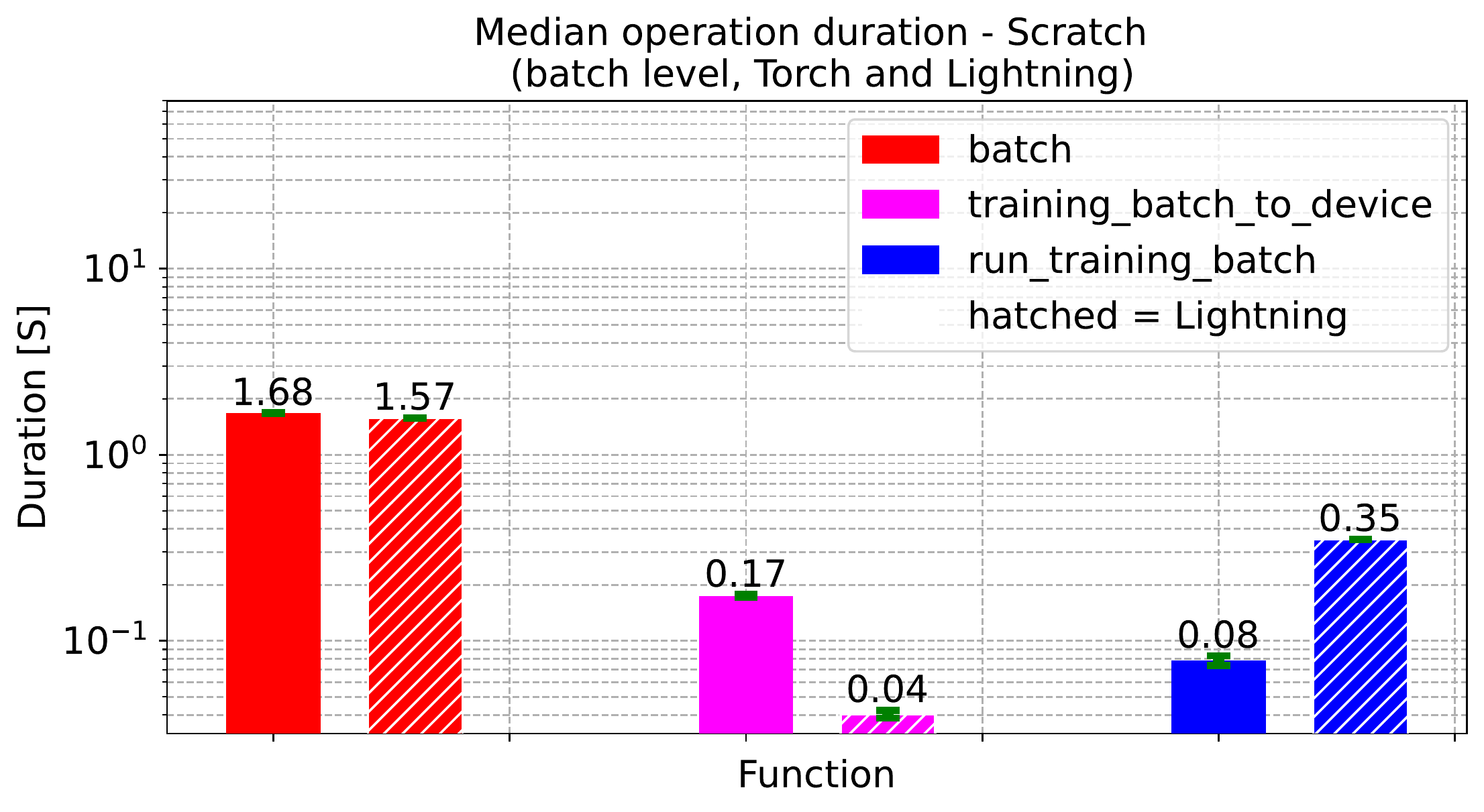} 
	\end{minipage}\hfill
	\begin{minipage}{0.59\linewidth}
	    \centering
        \includegraphics[width=\textwidth, trim={0 0 0 0cm}, clip]{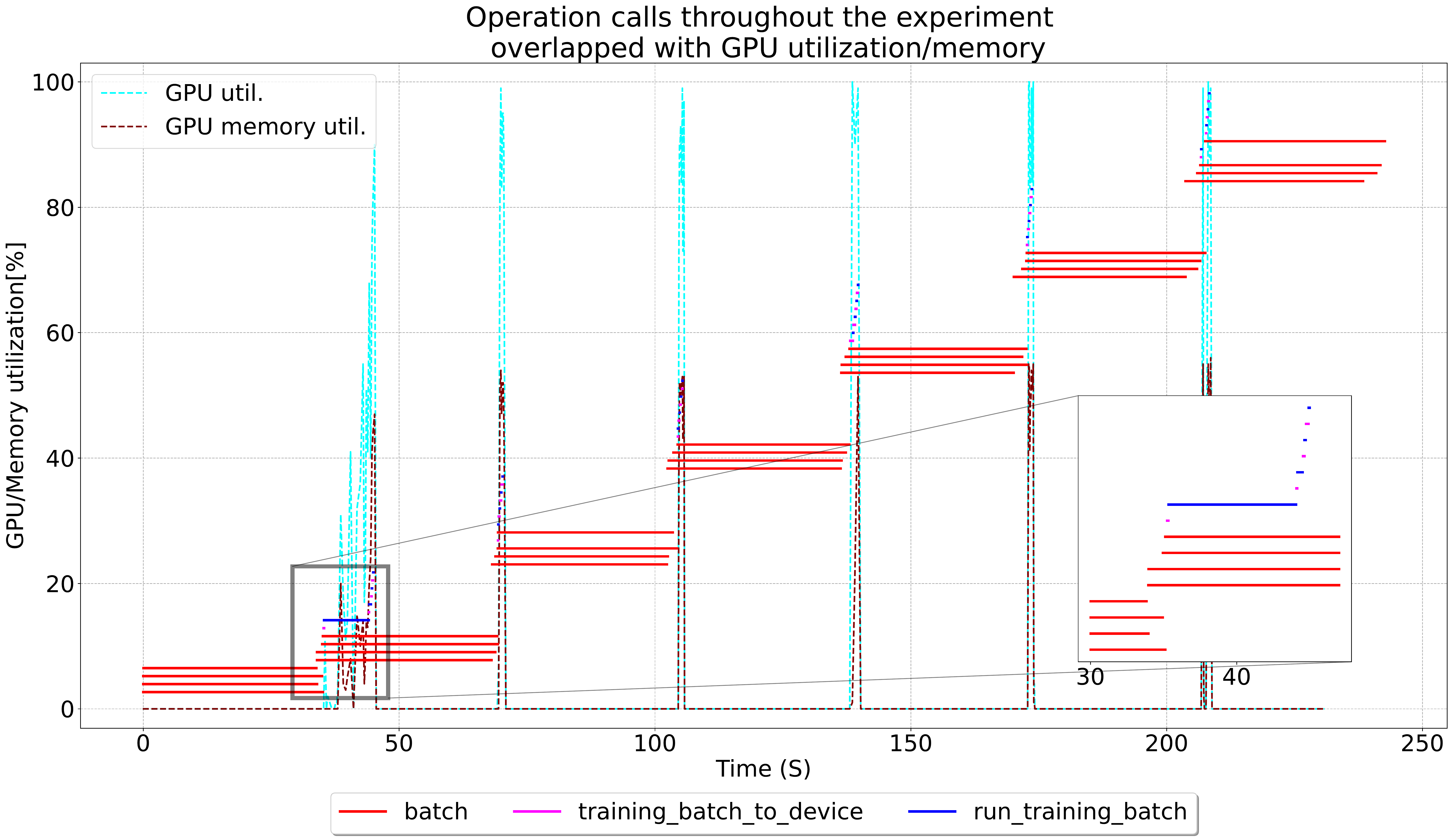} 
	\end{minipage}
    \vspace{2mm}
    \caption{Graphical representation of data loading function calls; left image shows average time necessary to load a batch, transfer to a GPU and train, while the right image shows a timeline of function calls, and GPU utilization (zoomed in on the first \SI{250}{\second} of the experiment, with AWS S3 storage).}
    \label{fig:ch1/function-duration-averages}
\end{figure}

Furthermore, when using AWS S3 remote storage the GPU idle time, and with it the runtime of the entire experiment, increased due to network latency. 
Therefore, when interpreting the GPU utilization in \autoref{tab:intro_motivation_example} with respect to \autoref{fig:dataloader-stack-full}, one may conclude that a large portion of time is used on data handling and not for training.
An attempt to make this conjecture visible is shown in \autoref{fig:ch1/function-duration-averages} which shows the timeline of the first 250 \si{\second} of training.
The horizontal lines in red, magenta and blue show the duration of the respective function calls in \autoref{fig:dataloader-stack-full} for loading a batch (\textit{Get batch}), moving the data to the GPU (\textit{Training batch to device}) and training on it (\textit{Run training batch}).
The cyan and brown dashed lines correspond to GPU utilization.

The right plot in \autoref{fig:ch1/function-duration-averages} shows that the \textit{training} starts by downloading 4 batches in parallel (red lines).
Once complete, there is almost an indistinguishable magenta dot that represents copying the training batches to the GPU.
Finally, there is a short blue line that represents the training.
One may notice that simultaneously with the start of the magenta line the GPU utilization and memory increases.
Worth mentioning is that the first training loop is slightly longer than the remaining training iterations due to the initial resource initialization.
In summary, this shows how the majority of time is used for data loading. 

 \autoref{fig:ch1/function-duration-averages} highlights several important considerations: 
 \begin{enumerate}
     \item Most of the plot consists of red lines which represent the lifetime of a function in charge for loading the batch (\textit{Get batch}).
     \item We see discrete steps and GPU utilization peaks with long pauses in between.
 \end{enumerate}
 This indicates that it takes only a short period of time to process data on the GPU for training compared to the time that is needed to load it from remote storage.
 For both Torch and Lightning, loading a batch from a remote storage takes more than \SI{30}{\second}. Left two plots of the \autoref{fig:ch1/function-duration-averages} show details for scratch and remote storage, with both Torch and Lightning. While for local storage, the batch loading time still takes the most time (between \SI{1.68}{\second} and \SI{1.57}{\second} (median)), comparing to S3 its only a fraction of time. As we do not have unlimited memory and since we cannot have an unlimited number of concurrent connections, we try to exploit the available network bandwidth by introducing within-batch parallel download of samples.

In the next section, we introduce an additional layer of parallelization and provide new experiments to explore how it affects the end-to-end performance.

%% file: chapters/ch2_parallelism.tex
\section{Data loading modifications}

\subsection{Problem analysis: data loading in PyTorch}

\begin{figure}[ht!]
\includegraphics[width=0.5\textwidth]{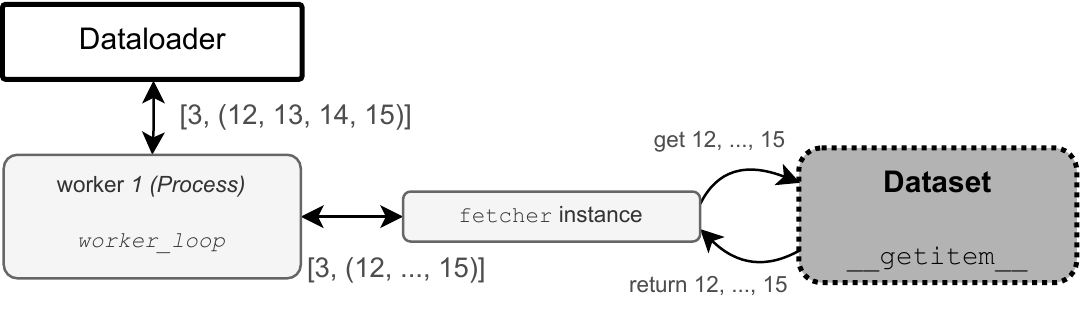} 
\centering
\caption{Simplified illustration showing sequential loading of data items.}
\label{fig:ch2/dataloader-index}
\end{figure}

While the data loading pipeline shown in \autoref{fig:dataloader-stack-full} uses parallelization over batches by utilizing multiple workers, the data items themselves are accessed sequentially within a batch\footnote{\url{https://github.com/pytorch/pytorch/blob/v1.9.1/torch/utils/data/_utils/fetch.py\#L26}}.
To visualize, \autoref{fig:ch2/dataloader-index} depicts the loading of a single batch which proceeds as follows.
The \pycode{Dataloader}\footnote{\url{https://github.com/pytorch/pytorch/blob/v1.9.1/torch/utils/data/dataloader.py\#L904}} uses a \pycode{worker\_loop}\footnote{\url{https://github.com/pytorch/pytorch/blob/v1.9.1/torch/utils/data/_utils/worker.py}} that gets attached to a new \pycode{Process}.
Also, the \pycode{Dataloader} creates an index queue for each worker and populates it with a tuple consisting of the batch ID and indices\footnote{\url{https://github.com/pytorch/pytorch/blob/v1.9.1/torch/utils/data/dataloader.py\#L1220 }}.
Those indices correspond to data items to be loaded.
To give an example, and entry $(3, [12, 13, 14, 15])$ would mean that the worker is loading batch number 3, consisting of four items $(12 ... 15)$.
The worker then creates a fetcher\footnote{\url{https://github.com/pytorch/pytorch/blob/v1.9.1/torch/utils/data/_utils/worker.py\#L268}}, which takes the list of indices to fetch and proceeds to do so sequentially with the \pycode{\_\_getitem\_\_} function of the \pycode{Dataset}.
Therefore, by using 4 workers at most 4 batches, are loaded in parallel, while the individual data items within a batch are loaded sequentially.
This raises the question: why are individual data items fetched sequentially?

\vspace{-2mm}
\subsection{Modifications: fetcher parallelization layer}
\vspace{-2mm}
In order to fetch individual data items in parallel, we propose adding an additional layer of concurrency, such that each worker can use multiple fetchers to get individual data items in parallel.
Considering the previous example, this would correspond to fetching the items $[12, ..., 15]$ in parallel. 

We equip \pycode{\_MapDatasetFetcher} (from \pycode{torch.data.\_utils}) with two new classes, each one using a different approach to parallelized downloads: Thread Pool and Asyncio. 
Threading is usually associated with parallelism using shared memory.
However, since Python uses the Global Interpreter Lock (GIL) to protect the internal state of the interpreter, all threads in Python are pinned to a single CPU, i.e. Process~\cite{hattingh2020using}.
The benefit of threading is in preemptive switching automatically performed by a thread manager.
However, this comes at a price of uncertainty since this switch may occur at any moment.
Therefore, to prevent incoherent states, it is required to use critical sections (enabled by locks), which can be expensive when using a large number of threads\footnote{The authors are aware that the term concurrency is associated with Asyncio, while parallelism is associated with Threads. In this work these terms are used interchangeably. }.

On the other hand, Asyncio allows for concurrency within a single thread.
By using the keywords yield and await task switching is performed manually.
This minimizes CPU utilization and has fewer overheads than threads.
The synchronization points are therefore no longer necessary.
However, this requires a non-blocking version, or async version, of operations such as file reading, networking, etc.\footnote{\url{https://pybay.com/site_media/slides/raymond2017-keynote/intro.html}}.  

Since both approaches have advantages and disadvantages with respect to this particular use case, the decision is to implement the existing \pycode{\_MapDatasetFetcher} with both libraries. 

\vspace*{-\baselineskip}
\begin{itemize}
    \setlength\itemsep{0em}
    \item \pycode{\_AsyncMapDatasetFetcher} is implemented using the \pycode{Asyncio} library.
    In this approach, \pycode{asyncio.Queue} is used for all items in the aforementioned batch index queue, and for each one creates a fetch task.
    Once this is completed, all tasks for fetching individual data items are started, and run asynchronously.
    When all the items are loaded, the batch is assembled and returned to the \pycode{Worker}, i.e. \pycode{Dataloader}.
    This implementation is referred to as \textit{Asyncio implementation} in the text.
    \item \pycode{\_ThreadedMapDatasetFetcher} has the same functionality as \pycode{\_AsyncMapDatasetFetcher} but uses the \pycode{Threading} library instead.
    Furthermore, it also introduces an additional layer of parallelism in the worker which is explained in the following paragraph.
    This implementation is referred to as \textit{Threaded implementation} in the text.
\end{itemize}

\vspace{-2mm}
\begin{figure}[hb!]
\includegraphics[width=0.89\textwidth,  trim={0 0 5mm 3.5mm}, clip]{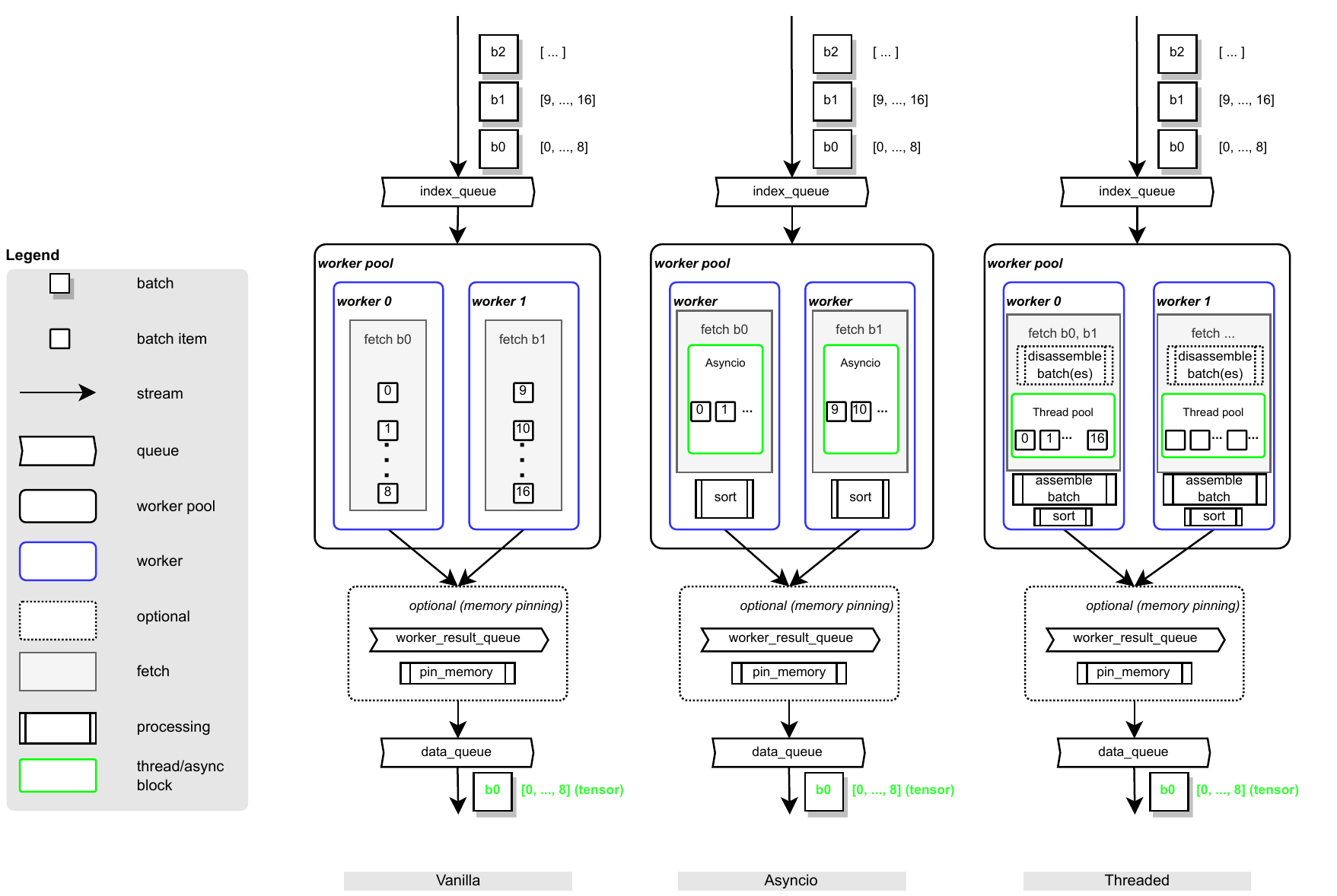} 
\centering
\caption{Implementation differences between the vanilla batch loading (left), and two new implementations introduced in this work; asyncio and threaded (respectively), }
\label{fig:ch2/parallelization-implementation-illustration}
\end{figure}

To highlight the differences between the different implementations, consider \autoref{fig:ch2/parallelization-implementation-illustration}.
In the illustrated examples two workers are used for loading the data items.
The left-most part of the image shows the vanilla implementation of the batch loading in PyTorch.
It allows for batch level parallelism, i.e. each worker processes a single batch at a time.
The \pycode{number\_of\_workers} and the \pycode{prefetch\_factor} determine the maximum number of batches loaded to memory and ready to be used for training.
The items of the batch 0 (b0) are loaded into the \pycode{index\_queue}, which is consumed by a worker (e.g. worker 0), which sequentially loads data items indexed $0, ...8$.
Optionally, if memory pinning is used the items are transferred to the pinned memory.
At this point the batch is ready for the trainer to consume.

The middle part of \autoref{fig:ch2/parallelization-implementation-illustration} illustrates the Asyncio implementation.
At first, a list of indices for the individual data items within a batch are passed from the worker to the fetcher.
However, instead of sequentially loading each data element, an Asyncio asynchronous block, which allows for concurrent execution, is added.
Since there is again no guarantee that the data items are loaded in the requested order, the elements are sorted and put to the \pycode{data\_queue} after the batch is complete.

Last but not least, the Threaded implementation is illustrated in the right part of  \autoref{fig:ch2/parallelization-implementation-illustration}.
It has one additional feature, which we tried out only in this impelementation.
In addition to the within-batch parallelism of Asyncio, we tried out batch disassembly in this implementation as well.
Instead of loading a single batch the worker disassembles several batches into individual data items and proceeds to download them.
The number of disassembled batches is controlled by the \pycode{batch\_pool} parameter.
If it is set to a value larger then 0 batches are disassembled, otherwise no batch pool is used.
The use-case illustrated here, has \pycode{batch\_pool} set to 16 which leads to two batches disassembled.
After disassembling, a \pycode{Thread} pool is used to fetch all 16 items in parallel.
As the data items are loaded, the batches are reassembled and their individual items are sorted to restore the requested order\footnote{due to parallelism, there is no guarantee elements are downloaded in a specific order}.
Finally, the batches are put into the \pycode{data\_queue} to be ready for the trainer to consume them.

Both Asyncio and Threaded are intuitive to use, and almost invisible to the user. While constructing the \pycode{Dataloader}, users can specify two new parameters, along with the implementation choice; \textit{Threaded}, \textit{Asyncio} and \textit{Vanilla}:

\vspace*{-\baselineskip}
\begin{itemize}
    \setlength\itemsep{0em}
    \item \textit{Number of fetch workers}, refers to the maximum number of threads the \pycode{ThreadPoolExecutor} can use to execute fetch tasks asynchronously.
    \item \textit{Batch pool size}, referring to the aforementioned batch pool, i.e. the number of items that each worker can take from the given batches.
\end{itemize}

\autoref{tab:ch2-parameters} summarizes the dataloading features and their mapping to the technical parameters. 

\begin{table}[ht!]
\centering
\resizebox{0.99\textwidth}{!}{
\begin{tabular}{@{}lllll@{}}
\toprule
\textbf{Feature} &
  \textbf{Description} &
  \textbf{vanilla} &
  \textbf{asyncio} &
  \textbf{threaded} \\ \midrule
\begin{tabular}[c]{@{}l@{}}parallelism over\\ batches\end{tabular} &
  \begin{tabular}[c]{@{}l@{}}The number of \\ batches that can be \\ downloaded concurrently\end{tabular} &
  $num\_workers$ &
  $num\_workers$ &
  $num\_workers \times \frac{batch\_pool}{batch\_size}$ \\ \midrule
Batch queue size &
  \begin{tabular}[c]{@{}l@{}}\textit{Backpressure}, i.e. the number of \\ batches that needs to be loaded \\ for training before a new batch \\ is fetched\end{tabular} &
  $num\_workers \times prefetch\_factor$ &
  $num\_workers \times prefetch\_factor$ &
  $num\_workers \times prefetch\_factor$ \\ \midrule
\begin{tabular}[c]{@{}l@{}}Batch item\\  parallelism\end{tabular} &  \begin{tabular}[c]{@{}l@{}}The number of concurrent \\ tasks that load single data items. \end{tabular} &
  -- (semantically 1) &
  $num\_fetch\_workers$ &
  $num\_fetch\_workers$ \\ \midrule
\begin{tabular}[c]{@{}l@{}} Batch disassembly\end{tabular} &
  \begin{tabular}[c]{@{}l@{}} The number of data items \\ loaded concurrently, across multiple\\ batches (as batches are disassembled).\end{tabular} &
  -- (semantically 1) &
  -- (semantically 1) &
  $batch\_pool$ \\ \bottomrule
\end{tabular}
}
\caption{Feature to technical parameter mapping}
\label{tab:ch2-parameters}
\end{table}

\subsection{Results}

For the proposed modifications, we perform a benchmark to compare how do Asyncio and Threaded implementation perform against the Vanilla implementation. We reuse the same parameters as for the motivational experiment (\autoref{tab:motivational-experiment-params}), with several additions shown in \autoref{tab:parallelization-params}.

\begin{table}[ht!]
\centering
\resizebox{0.65\textwidth}{!}{
\begin{tabular}{@{}lllllllll@{}}
\toprule
\textbf{\begin{tabular}[c]{@{}l@{}}Batch \\ size\end{tabular}} &
  \textbf{Workers} &
  \textbf{\begin{tabular}[c]{@{}l@{}}Prefetch\\ factor\end{tabular}} &
  \textbf{\begin{tabular}[c]{@{}l@{}}Number of \\ fetchers\end{tabular}} &
  \textbf{\begin{tabular}[c]{@{}l@{}}Batch\\ pool\end{tabular}} &
  \textbf{\begin{tabular}[c]{@{}l@{}}Dataset\\ limit (size)\end{tabular}} &
  \textbf{\begin{tabular}[c]{@{}l@{}}Learning\\ rate\end{tabular}} &
  \textbf{\begin{tabular}[c]{@{}l@{}}Weight\\ decay\end{tabular}} &
  \textbf{Epochs} \\ \midrule
\multicolumn{1}{c}{256} &
  \multicolumn{1}{c}{4} &
  \multicolumn{1}{c}{4} &
  \multicolumn{1}{c}{16} &
  \multicolumn{1}{c}{0} &
  \multicolumn{1}{c}{15000} &
  \multicolumn{1}{c}{0.1} &
  \multicolumn{1}{c}{0.0001} &
  \multicolumn{1}{c}{5} \\ \bottomrule
\end{tabular}
}
\caption{Parameters parallelization benchmarking}
\label{tab:parallelization-params}
\end{table}

\begin{figure}[ht!]
\includegraphics[width=0.75\textwidth]{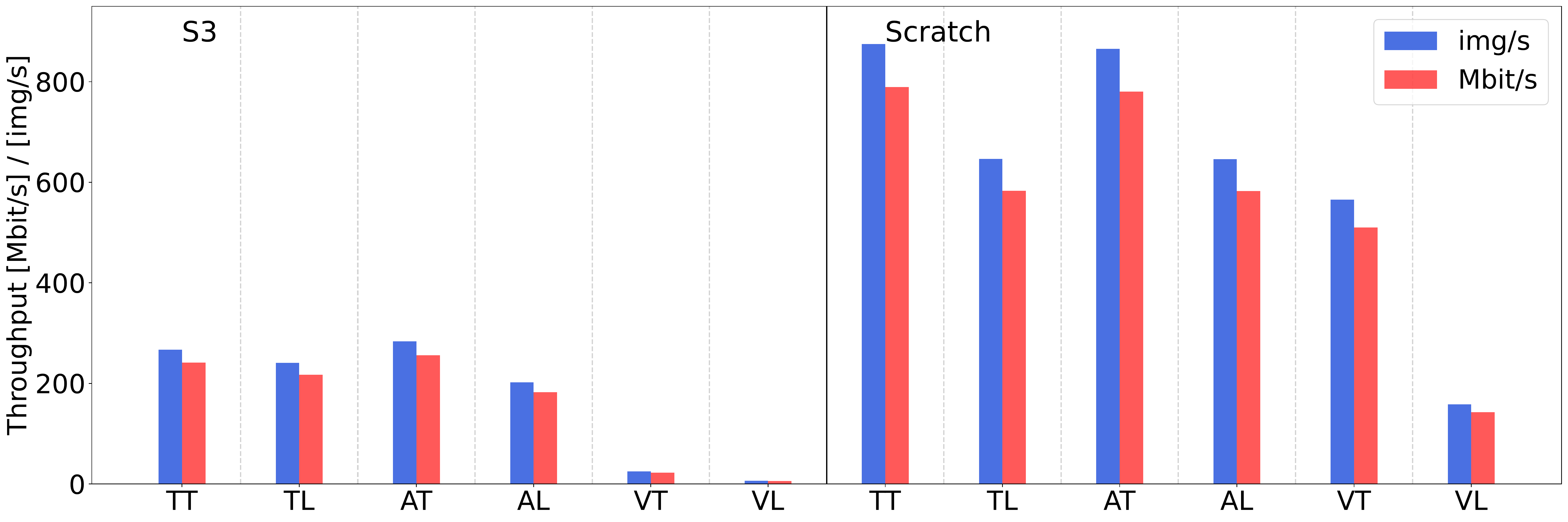}
\centering
\caption{Benchmarking data loading without and with our modifications for s3 and scratch. The $y$-axis shows the throughput (\si{\mega\bit\per\second} and \si{\img\per\second}) for the newly added parallelism in the fetching layer. The $x$-axis shows abbreviations for libraries and implementations (e.g. \textit{TT} is PyTorch Threaded, \textit{TL} is PyTorch Lightning, \textit{AT} is Asyncio Threaded, etc.). Parameters according to Table~\ref{tab:parallelization-params}.}
\label{fig-tab:ch2/fetch-parallelized}
\end{figure}

\vspace{-5mm}
The experiment results are shown in \autoref{fig-tab:ch2/fetch-parallelized}. The left side of the figure shows the results for the cloud based storage (S3), while the right part shows the results for local storage. 
The throughput for loading the data from S3 storage using the Asyncio and Threaded implementation compared to the Vanilla implementation improved by $11.44$ and $10.77 \times$, respectively, when used for the training setup with pure PyTorch.
For the training setup with PyTorch Lightning the improvements are $32.94$ and $39.20 \times$ respectively.
When loading the data from Scratch storage the gain in throughput for both, the Asyncio and Threading implementation, is $1.55 \times$ for the training setup with pure PyTorch  and $4.07 \times$ for the PyTorch Lightning setup.

The results show that cloud based storage is benefiting the most from the introduced layer of parallelism.
This is expected, as there is substantial latency when fetching from cloud storage (see \autoref{dataset}), which is particularly bad under sequential access pattern. This implicates that additional fine tuning should produce even better results, which is a pointer for further investigation. Even with this change, local storage is outperformed in the case Vanilla Lightning vs any of our modifications.

\begin{figure}[ht!]
\includegraphics[width=0.4\textwidth]{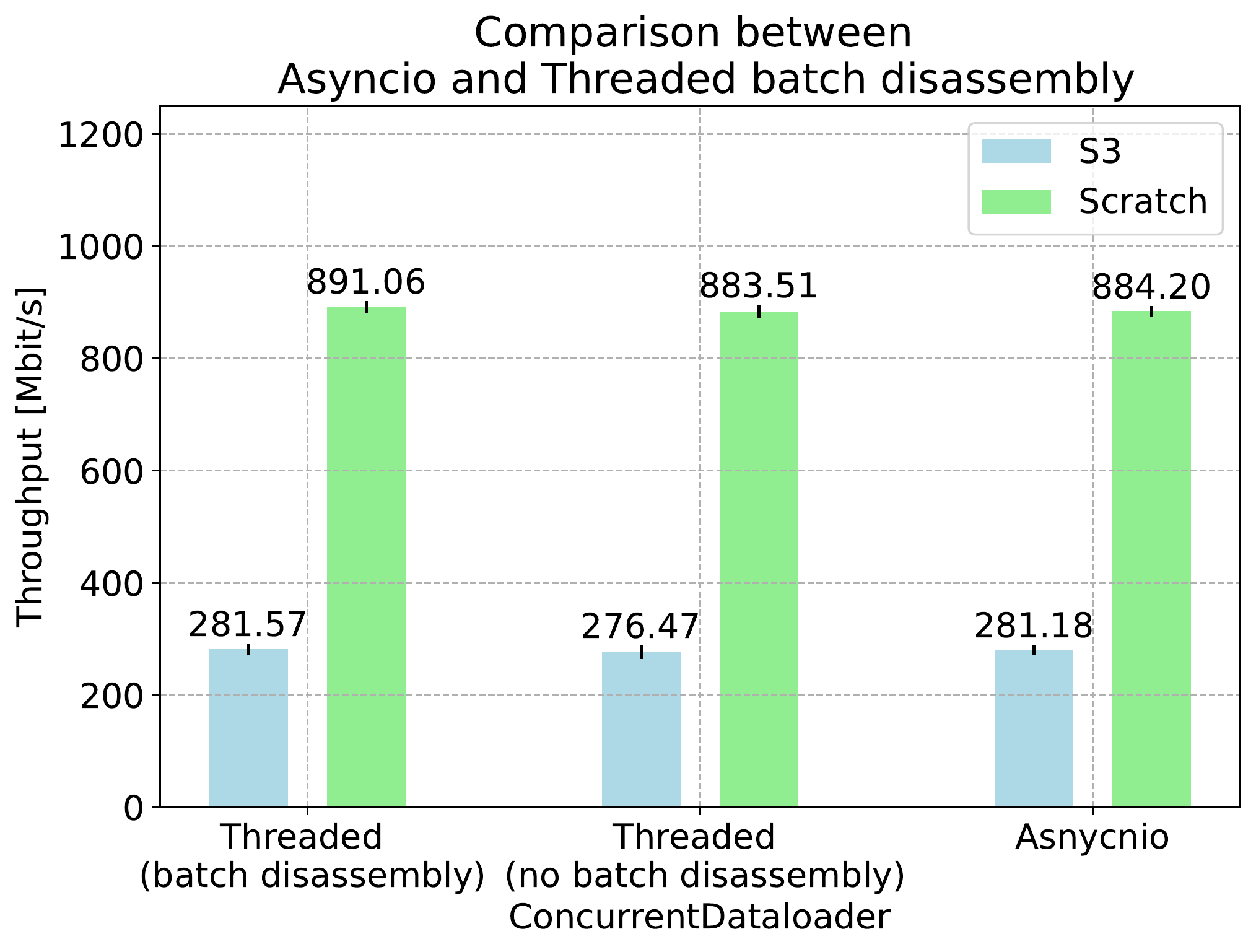}
\centering
\caption{Comparison between different approaches of data-item parallelism. Threaded with and without batch disassembly vs. Asyncio. }
\label{fig:ch2/threaded_vs_asnync}
\end{figure}

Since Threaded implementation can come in two different forms, as explained previously, with and without the feature of batch disassembly (\autoref{fig:ch2/parallelization-implementation-illustration}), we performed an additional benchmark comparing the three different approaches, with the results shown in \autoref{fig:ch2/threaded_vs_asnync}.

The proposed feature, which disassembles a batch and then downloads data items from multiple batches in parallel, within a single worker seems to provide no significant improvement, for the proposed use-case. Henceforth, this feature will not be used. 

\subsection{Additional considerations: {Process} initialization and caching}

Through various benchmarks performed for this work, we learned about two additional points that should be taken into consideration for a potential performance increase, \pycode{Process} creation and \pycode{Dataloader} initialization. Also, we look into the effect of using a web cache.

\subsubsection*{\pycode{Process} creation}

{Process} creation, to initialize worker \pycode{Process(es)}, both PyTorch Lightning and PyTorch use Python's \pycode{Multiprocessing} library.
However, for PyTorch the default setting is \pycode{fork}, whereas for PyTorch Lightning it is the \pycode{spawn} method\footnote{\url{https://docs.python.org/3.10/library/multiprocessing.html\#contexts-and-start-methods}}.
The main difference between these two methods is that \pycode{fork} creates a child process, which inherits all resources from the parent, and that \pycode{spawn} creates a clean, new Python interpreter process, which takes significantly more time.
This also means that when using the \pycode{fork} method, CPU and GPU calls cannot be mixed, which in PyTorch and PyTorch Lightning means that \textit{memory pinning} cannot be used. 
This is due to the fact that memory pinning uses GPU calls to automatically load fetched data into page-locked (i.e. pinned) memory, resulting in faster host to GPU copying\footnote{\url{https://pytorch.org/docs/stable/data.html\#memory-pinning}}.

\begin{figure}[ht!]
\includegraphics[width=0.65\textwidth]{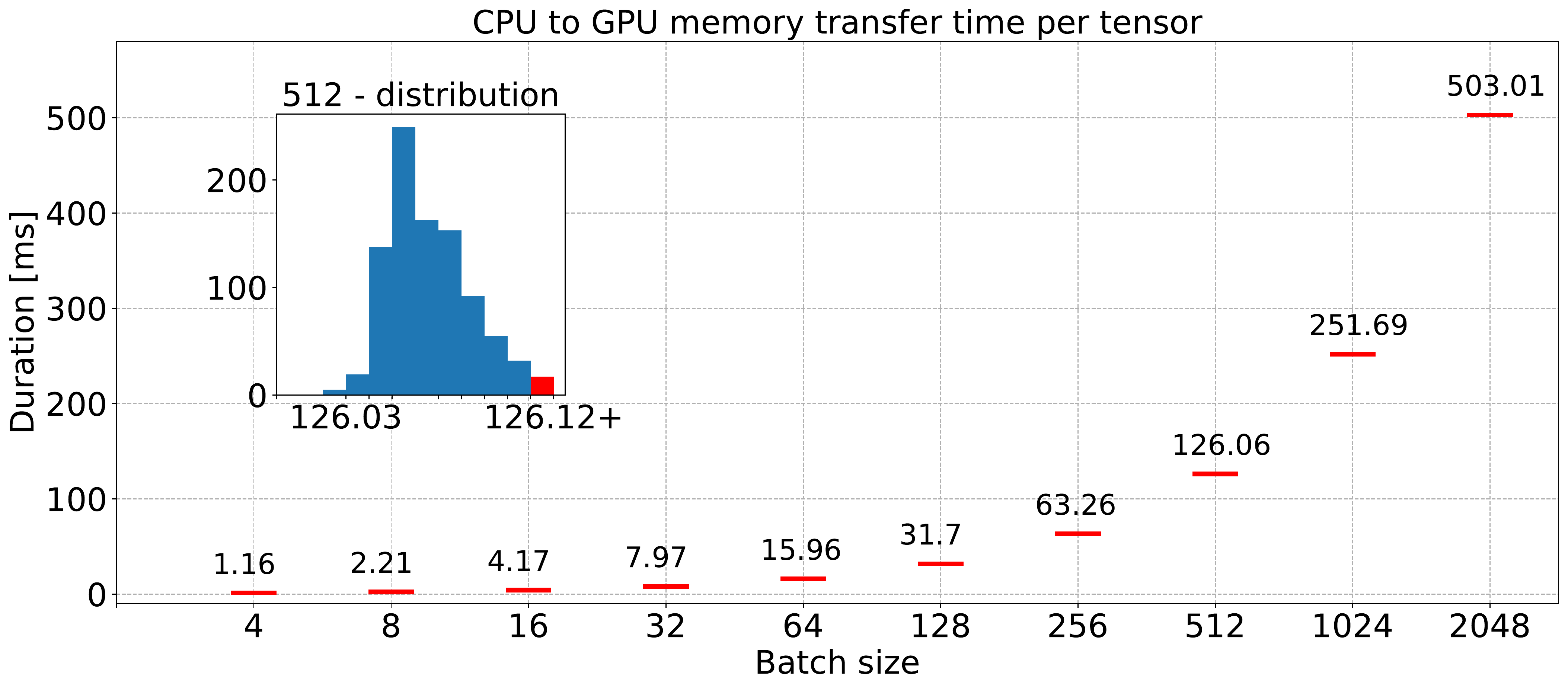}
\centering
\caption{Increasing the batch size, increases the average CPU to GPU tensor transfer time. The inset plot shows the distribution histogram of transfer time for batch size 512, with the red bar representing the overflow bin that contains all outliers}
\label{fig:ch2/image-loading}
\end{figure}

\autoref{fig:ch2/image-loading} shows that larger batch size significantly increase the CPU to GPU transfer time, so memory pinning should be an important feature. However, for our use-case the GPU transfer time (training batch to device, shown in \autoref{fig:ch1/function-duration-averages}) is not significant comparing to the batch loading time, which we are trying to reduce.

\subsubsection*{\pycode{Dataloader} initialization}

\begin{figure}[ht!]
\includegraphics[width=0.65\textwidth]{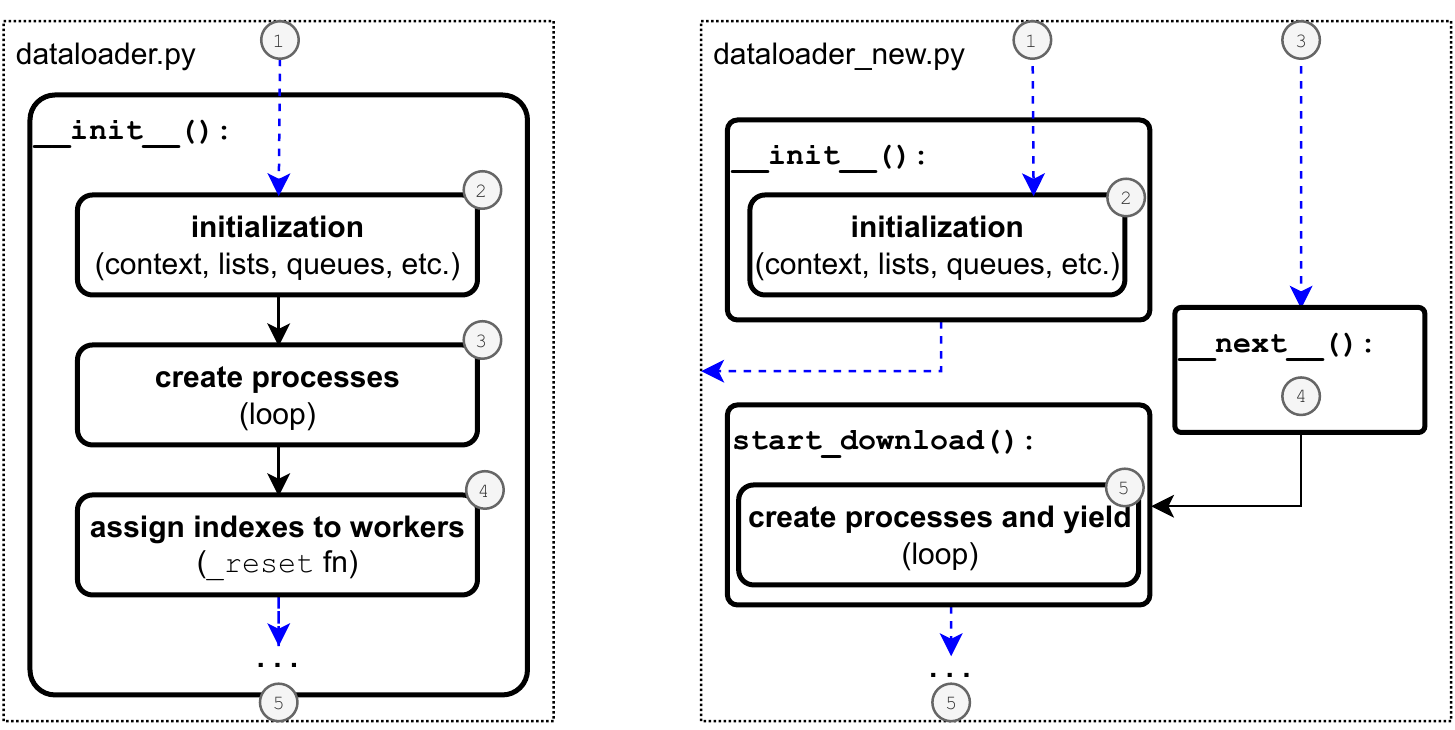} 
\centering
\caption{Simplified illustration of steps necessary to start the data loading. The left-hand side figure shows the original, while the right side shows a modified, non-blocking version with lazy initialization.}
\label{fig:ch2/dl-init}
\end{figure}

\pycode{Dataloader} initialization (constructor), creates \pycode{Process(es)} which are used as workers to load data items of a single batch.
The issue here is that creating \pycode{Process(es)} can be slow which further delays the object creation\footnote{\url{https://github.com/pytorch/pytorch/blob/v1.9.1/torch/utils/data/dataloader.py\#L904}} and execution.
It is not good practice to have bloated constructors since their role should be to construct objects.
However, often it is also used for initialization or obtaining the necessary dependencies.
In the case of the \pycode{Dataloader} constructor a red flag is the Process initialization, which may take considerable time and therefore block further execution.
Suppose that the \pycode{Dataloader} is constructed with 16 workers (i.e. \pycode{Process(es)}) and each one is taking a second to initialize.
This will take 16 seconds before the first process starts loading batches.
Though there can be multiple solutions to deal with this, we introduced a modification that allows for lazy-loading and non-blocking process creation which is shown in \autoref{fig:ch2/dl-init}. 

Blue dashed lines represent external calls coming to \pycode{dataloader.py}.
Normally, as shown in the left side of \autoref{fig:ch2/dl-init}, (1) a call comes in requesting the initialization of the \pycode{\_MultiProcessingDataLoaderIter}, an object in charge for loading data.
Initialization takes place by asserting worker sizes (2), creating a multiprocessing context, initializing local counters, flags, lists, etc.
Then (3), the code enters a process creation loop, which starts multiple new \pycode{Process(es)} (this is a parameter, selected by the user).
The loop sequentially creates new processes.
After completion, a \pycode{\_reset} function (5) is called, which among other things, calls the \pycode{\_try\_put\_index} function that is in charge for passing the index queues to individual workers.\\
We addressed the problem of the blocking loop by first utilizing lazy loading, and yielding the \pycode{Process(es)} that have already been created thereafter.
Similarly then described above, (1) the multiprocessing context, local flags, lists, etc. are initialized.
After this (2), the \pycode{\_init\_\_} function returns and allows for the \pycode{\_MultiProcessingDataLoaderIter} object to be created.
Thereafter, once the training loop requests a new batch (3), it uses the \pycode{\_\_next\_\_} function, which is the part of    \pycode{\_MultiProcessingDataLoaderIter} object to obtain the batch.
At this point, a new function (\pycode{start\_download}) is triggered (4), which creates new \pycode{Process(es)} (i.e. workers) within a loop (5).
Instead of blocking, it yields the created Process and lets it proceed with the data loading, but not before the call of the new \pycode{\_try\_put\_index} function which loads the necessary indexes, for the processes that have been created up to that point.
This procedure is triggered for each epoch. 

\subsubsection*{Caching}

During each epoch, the training loop iterates through all the elements in a dataset.
When considering cloud storage, this means downloading the same data over and over again.
This download uses HTTP requests to fetch individual data items which opens the question of cashing.
Caching would allow to keep local copies of items that have already been downloaded and reuse them.
Also, by using cashing libraries we can determine how many items we want to store locally. 

For this purpose we use Varnish\footnote{\url{https://www.varnish-software.com/wiki/}}, which is usually used as an HTTP accelerator for heavily consumed API endpoints.
It intercepts HTTP requests at the OS level and checks if the data is already available locally.
If this is the case (cache hit), it returns the cached item and otherwise (cache miss) downloads it.

\begin{figure}[ht!]
	\begin{minipage}{0.49\linewidth}
        \centering
        \includegraphics[width=\textwidth]{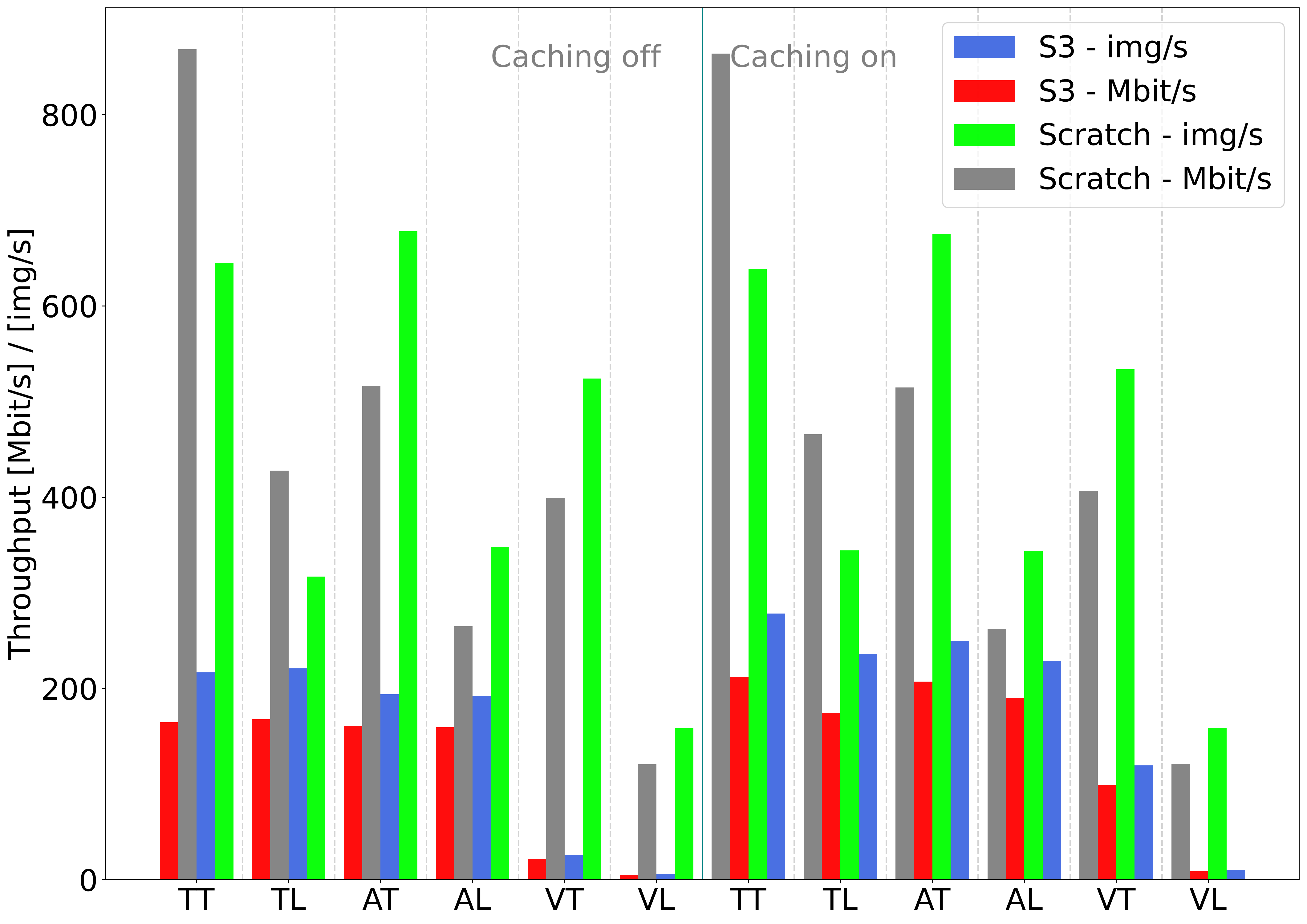} 
	\end{minipage}\hfill
	\begin{minipage}{0.49\linewidth}
	    \centering
        \includegraphics[width=\textwidth]{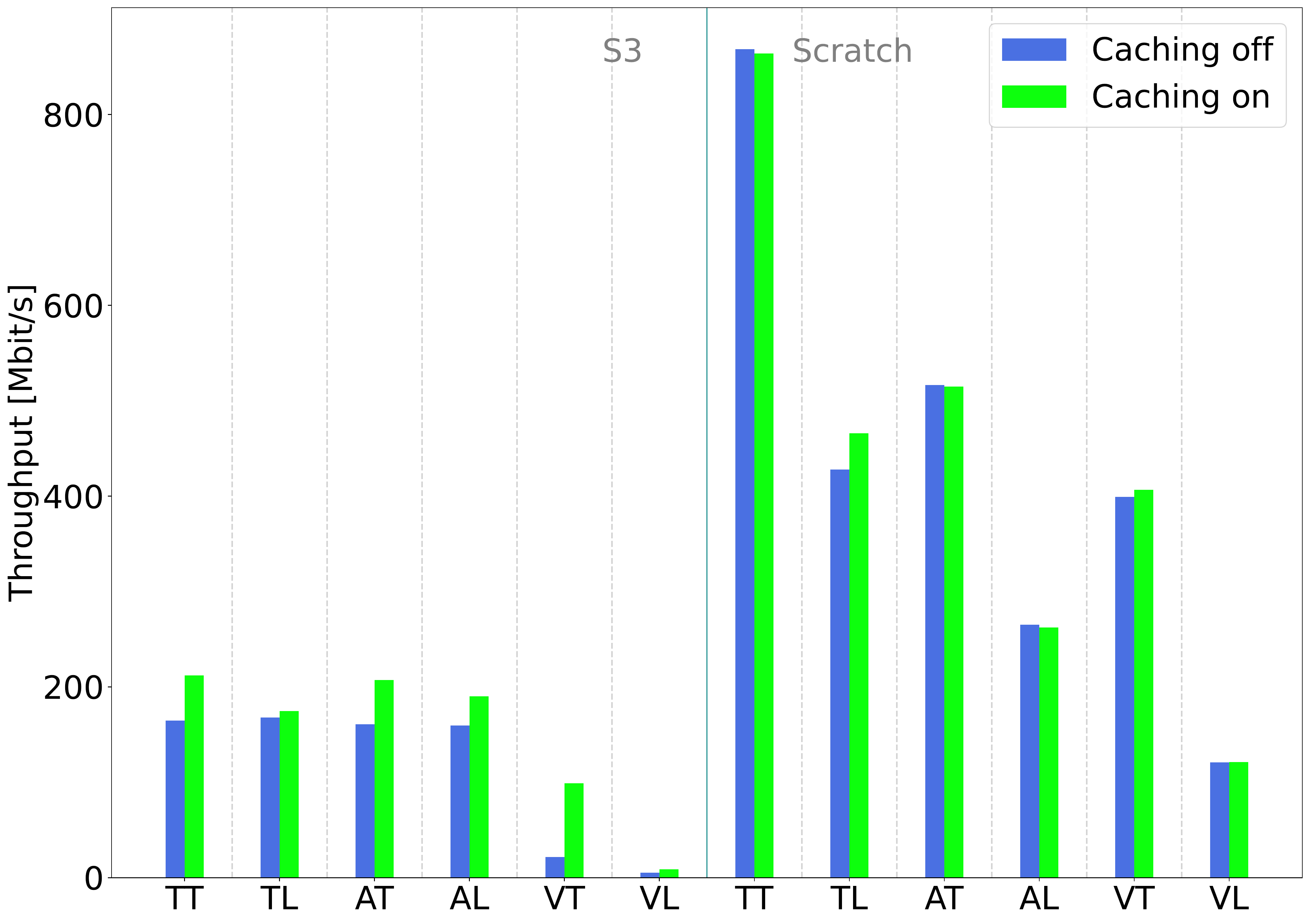} 
	\end{minipage}
    \vspace{2mm}
    \caption{The left figure shows the end-to-end throughput for both images and \si{\mega\bit\per\second}, with and without caching for both scratch and S3 storage. The $x$-axis shows abbreviations for libraries and implementations (e.g. \textit{TT} is Torch Threaded, \textit{TL} is Torch Lightning, \textit{AT} is Asyncio Threaded, etc.). The figure on the right highlights throughput differences in \si{\mega\bit\per\second} for S3 and scratch.}
    \label{fig:ch2/cashing}
\end{figure}

Using the same parameters as for the previous experiment (shown in \autoref{tab:parallelization-params}), \autoref{fig:ch2/cashing} shows the benchmark results with and without cashing.
To restrict caching to the use case that is studied in this work, which is that local storage is not sufficient to hold the full dataset, the caching size is set to \SI{2}{\giga\byte}.

For S3, the most significant improvements were found for Threaded Torch (\SI{28}{\percent}) and Vanilla Torch (\SI{450}{\percent}), and no improvements for the others (e.g. Threaded Lightning, Vanilla Lightning).
However, these results need to be taken with a grain of salt.
Since the caching size is limited the cache lookup often results in a cache miss, which requires the requested item to be downloaded again.
This is due to the fact that during each training iteration the access pattern to the items is random.
As a sanity check we also use caching for local storage and found that there are no significant differences as one would expect.
This leaves cache as an open question and a choice, depending on a particular use-case.

%% file: chapters/ch3_dataset_dataloader.tex
\section{Concurrency parameters for the \texttt{Dataset} and the \texttt{Dataloader}}

We demonstrated significant improvements in training runtime and throughput by introducing a new layer of concurrency.
Furthermore, we improved the \pycode{Dataloader} initialization process and addressed caching.
However, there is still a gap in performance between S3 and local storage that may be investigated further.
In this section, we study the performance gap between PyTorch and PyTorch Lightning, which is substantial.
In addition, we address how to select the concurrency parameters to squeeze the most of our hardware.
So far, we have observed that a large portion of the experiment runtime is used for data loading and not for actual training.
Therefore, we focus on the performance of the \pycode{Dataloader} (\textit{Get batch} in \autoref{fig:dataloader-stack-full}), and the \pycode{Dataset} (\textit{Get item} in \autoref{fig:dataloader-stack-full}) in the following experiments.  

\subsection{\texttt{Dataloader}}

By introducing a new concurrency layer, new parameters have been introduced.
However, the most important and influential ones are the number of fetch workers in combination with the number of workers.
Those two parameters define how many parallel Processes or Threads are used.
This can have a major impact on possible bottlenecks that may arise from resource locking.
The effect of these two parameters on the throughput is measured with the following experiment setting.

\begin{table}[ht!]
\centering
\resizebox{0.5\textwidth}{!}{
\begin{tabular}{@{}lllll@{}}
\toprule
\textbf{\begin{tabular}[c]{@{}l@{}}Batch\\ size\end{tabular}} &
  \textbf{\begin{tabular}[c]{@{}l@{}}Number of\\ batches\end{tabular}} &
  \textbf{Storage} &
  \textbf{\begin{tabular}[c]{@{}l@{}}Number of \\ workers\end{tabular}} &
  \textbf{\begin{tabular}[c]{@{}l@{}}Number of \\ fetchers\end{tabular}} \\ \midrule
\multicolumn{1}{c}{64} &
  \multicolumn{1}{c}{40} &
  \multicolumn{1}{c}{S3, scratch} &
  \multicolumn{1}{c}{1, 2, 4, 8, 16, 32, 64, 128} &
  \multicolumn{1}{c}{1, 2, 4, 8, 16, 32} \\ \bottomrule
\end{tabular}
}
\caption{Dataloader benchmark parameters}
\label{tab:dataloader-params}
\end{table}

\begin{figure}[ht!]
\includegraphics[width=0.8\textwidth]{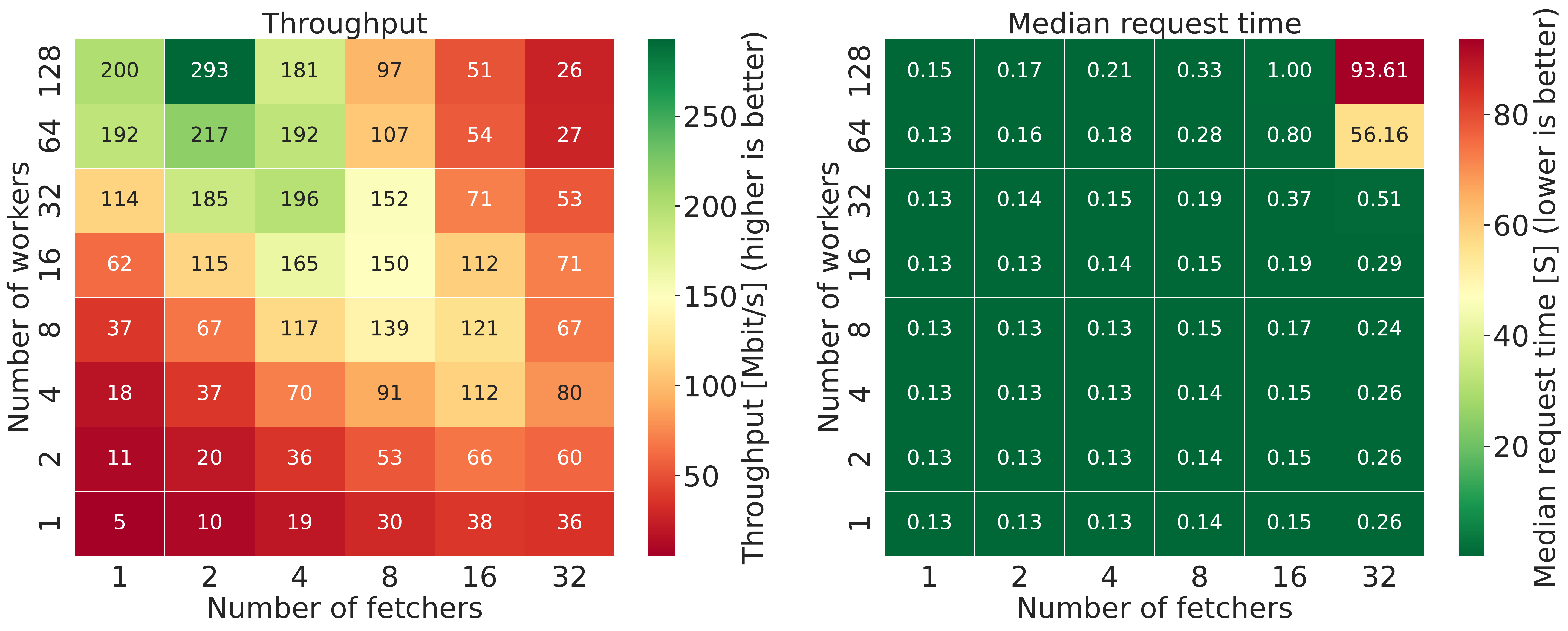}
\centering
\caption{Heatmap plot, showing how the number of workers and fetchers with S3 storage influence the throughput (left), and the response time (right), using the Threaded implementation.}
\label{fig:ch3/s3_throughput}
\end{figure}

\autoref{fig:ch3/s3_throughput} shows how increasing the number of fetchers and workers using S3 storage and Threaded implementation, influences the overall throughput in \si{\mega\bit\per\second} (left), as well as the median request time in \si{\second} (right).
Considering the left heatmap, we may notice that the best performance is achieved when $1-4$ fetchers and $32-128$ workers are used.
It also shows that a high number of workers and fetchers, as well as low numbers, result in suboptimal performance.
In the right heatmap, we may observe that the median request time is highest when we have many workers and fetchers, which may be related to resource locking.

\begin{figure}[ht!]
\includegraphics[width=0.8\textwidth]{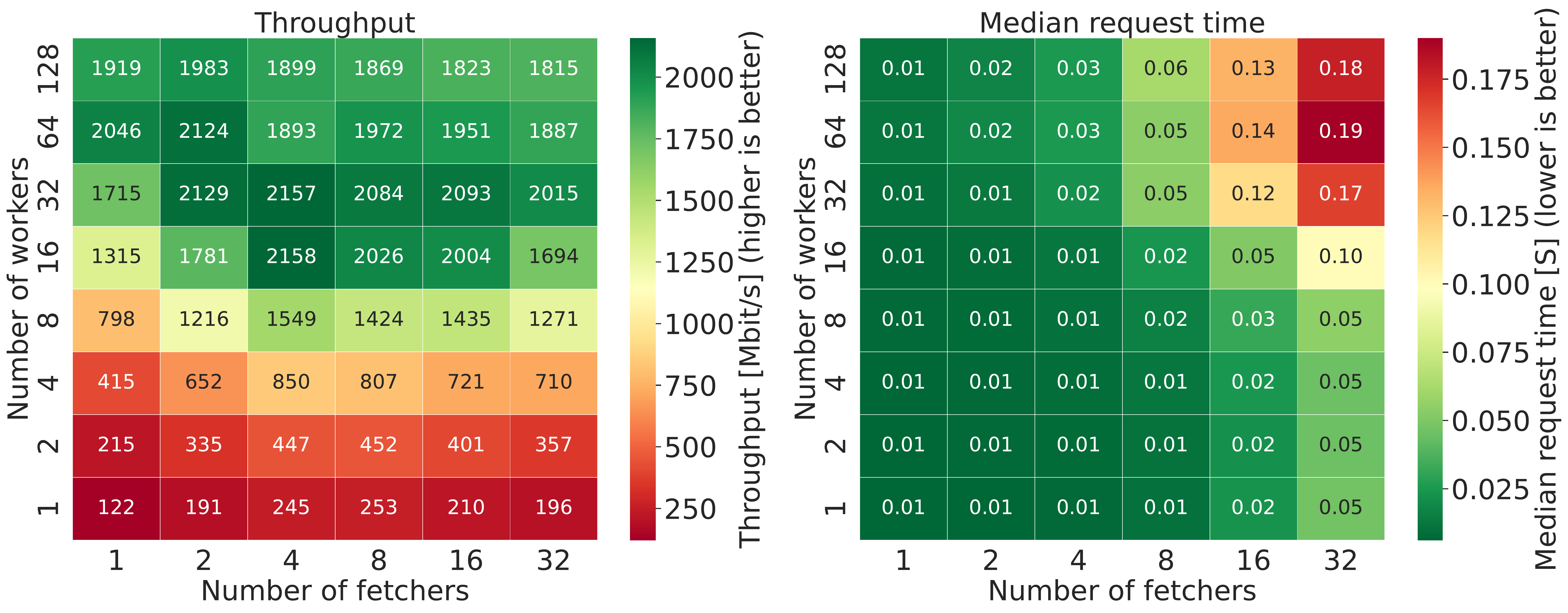}
\centering
\caption{Heatmap plot, showing how the number of workers and fetchers with scratch storage influence the throughput (left), and the response time (right), using the Threaded implementation (throughput values are rounded to the nearest decimal)}
\label{fig:ch3/scratch_throughput}
\end{figure}

\autoref{fig:ch3/scratch_throughput} shows the same analysis but for scratch storage. 
Besides the much higher throughput, we may notice that for median request time (right plot), we have similar deterioration in performance for a large number of workers and fetchers.
However, the throughput seems to be more stable with respect to the number of fetchers compared to S3 storage.
As a result, choosing the number of workers higher than $16$ leads to decent results, with the best throughput being achieved for $2-8$ fetchers.
This leads to the conclusion that the benefit from fetching is much less pronounced, since the median request time is already low and the networking overhead of remote storage is not present.  


\subsection{\texttt{Dataset}}
\label{dataset}

A layer below the \pycode{Dataloader} is the \pycode{Dataset}(\autoref{fig:dataloader-stack-full}) which directly loads the images from storage.
At this level, there are no workers or fetchers.
Therefore, it can be instantiated separately, since it is almost completely isolated from the ML framework.
Considering that this layer is accessing individual data items, it is important to understand how concurrency influences its performance.
In this section, we address the concurrency in the context of pure image loading using the \pycode{Dataset} instance and by increasing the multiprocessing pool size.
As the \pycode{Dataset} usually performs data augmentation, we did not exclude it from this benchmark. 
For this purpose, after the image is loaded from storage we perform 1) random resized crop to the dimension of 224x224, 2) horizontal flip, 3) conversion to tensor, and 4) normalization.
The parameters used for the experiment are the following:

\begin{table}[ht!]
\centering
\resizebox{0.8\textwidth}{!}{
\begin{tabular}{@{}lllll@{}}
\toprule
\textbf{\begin{tabular}[c]{@{}l@{}}Pool size\end{tabular}} &
  \textbf{\begin{tabular}[c]{@{}l@{}}Random images \\ loaded \end{tabular}} &
  \textbf{\begin{tabular}[c]{@{}l@{}}Number of image \\ groups (batches)\end{tabular}} &
  \textbf{\begin{tabular}[c]{@{}l@{}}Storage\end{tabular}} \\ \midrule
\multicolumn{1}{c}{1, 2, 3, 4, 5, 6, 7, 10, 15, 20, 30, 40, 50, 60, 80} &
  \multicolumn{1}{c}{2000} &
  \multicolumn{1}{c}{40} &
  \multicolumn{1}{c}{S3, scratch} \\ \bottomrule
\end{tabular}
}
\caption{Dataset benchmark parameters}
\label{tab:dataset-params}
\end{table}

In the experiment, 40 image batches (i.e. batches, but not to be confused with batches from the context of a Dataloader, as Dataloader is not used here) loaded using the pure \pycode{Dataset} object (without the upper layers of the data loading pipeline \autoref{fig:dataloader-stack-full}), each accessing 2000 random images while increasing the multiprocessing pool, that refers to the maximum number of parallelized functions with Python's \pycode{multiprocessing.Pool}\footnote{\url{https://docs.python.org/3/library/multiprocessing.html}}.
To randomly load an image, we've added a \pycode{get\_random\_item} function that randomly generates an integer index of an image, and then uses the usual \pycode{\_\_getitem\_\_} function to fetch the image from S3 or scratch.
This also means that the same data augmentation is performed, as in all other experiments. 

\begin{figure}[ht!]
	\begin{minipage}{0.49\linewidth}
        \centering
        \includegraphics[width=\textwidth]{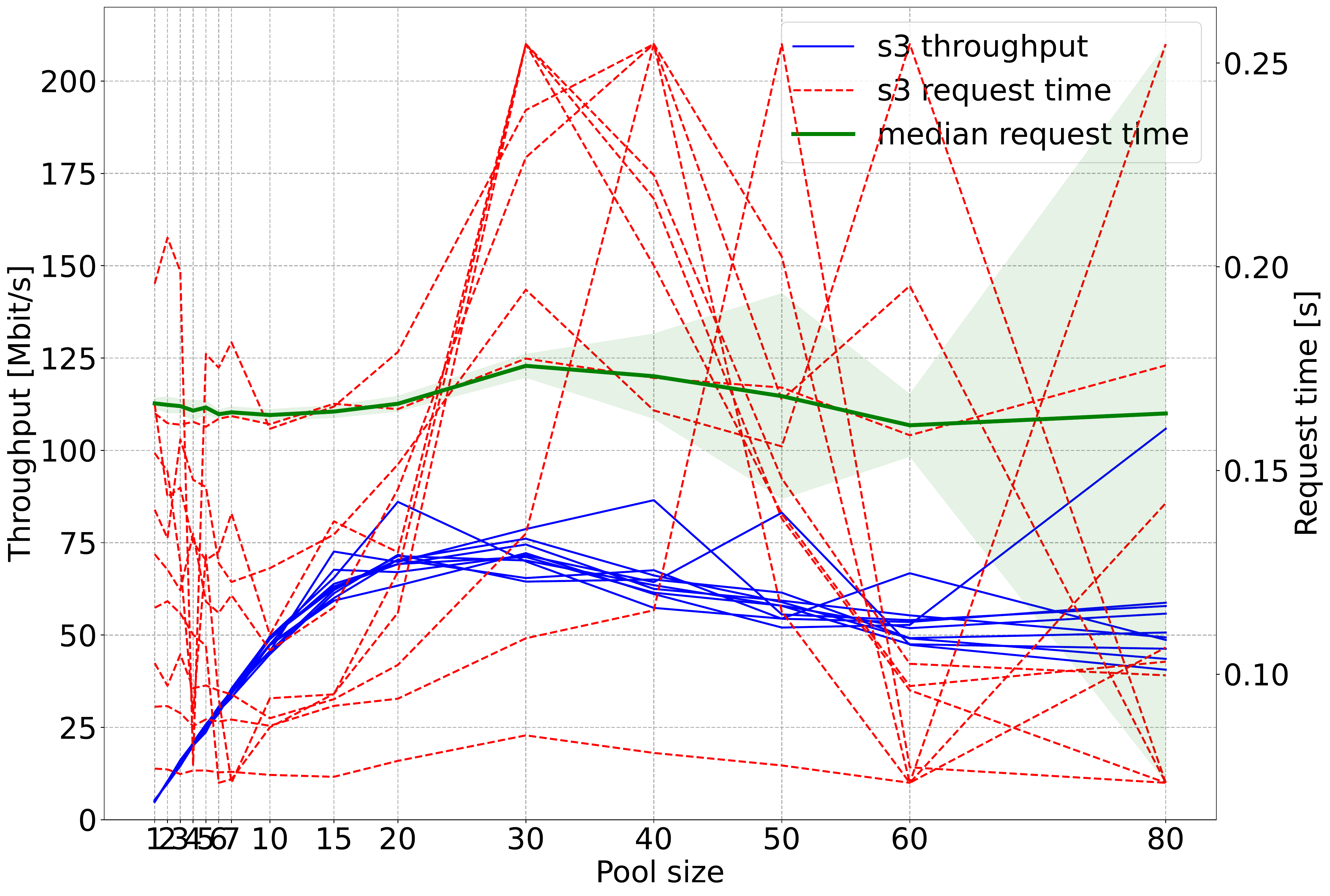} 
	\end{minipage}\hfill
	\begin{minipage}{0.49\linewidth}
	    \centering
        \includegraphics[width=\textwidth]{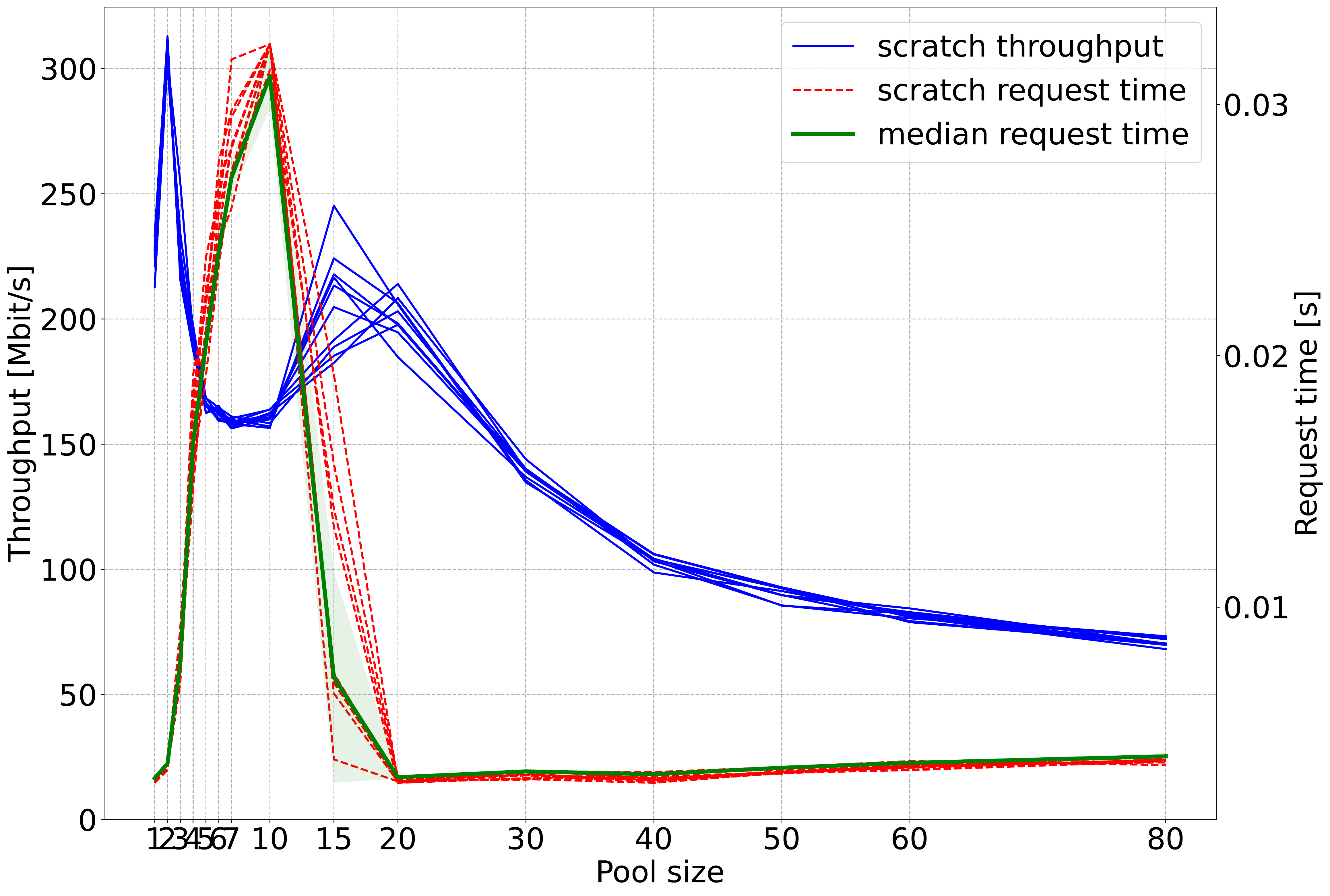} 
	\end{minipage}
    \vspace{2mm}
    \caption{Throughput (left $y$-axis) and request time (right $y$-axis) for random image loading using a \pycode{Dataset} object, for different multiprocessing pool sizes, for S3 (left) and scratch (right). Points of the blue line, represent throughput over the entire experiment, while the request time represents a median response time.}
    \label{fig:ch3/dataset-throughput}
\end{figure}

\autoref{fig:ch3/dataset-throughput} shows experiment result, i.e. the throughput and response time, for the increasing multiprocessing pool size, for both S3 and scratch, repeated 10 times.
On both plots, the blue lines represent the achieved throughput, for a certain multiprocessing pool size.
For S3, we can notice that after 30 simultaneous processes, there is almost no improvement in the throughput, which peaks around \SI{75}{\mega\bit\per\second}.
This is very similar to the previous case with the \pycode{Dataloader} (\autoref{fig:ch3/s3_throughput}) where we had the highest throughput for 32 workers and above, but then each worker (i.e. a process) had additional parallelized fetchers, which here is not the case.

In this experiment, using only the Dataset, the concurrency from multiple fetchers and workers is not present, and the only concurrency is on the image loading level, i.e. accessing random images. In terms of the Dataloader this corresponds to having no workers but multiple fetchers (in this case, pool size).
The maximum we are getting here, represents a maximum we can get per parallelized fetcher, i.e. roughly \SI{75}{\mega\bit\per\second}.
However, the Dataloader also uses Processes to represent workers, so, for instance with 4 workers, we could technically get \SI{300}{\mega\bit\per\second}. 

The red dashed lines show the request time which is the time necessary to read and return a single data item (with data augmentation included). Given that values are ranging from \SI{0.01}{\second} up to \SI{0.43}{\second}, it is hard to conclude whether the request time depends on networking parameters, number of threads, or both.
However, the median value suggests that between 20 and 60 workers in the pool, the response time is the highest.
Furthermore, comparing to measurement with respect to Scratch storage on the right, it does suggest that networking introduces unpredictable behavior with respect to the request time.
This might be the result of network latency, load, hardware, routing, etc. 

For scratch, the results are much different, showing two distinct throughput peaks, one for 2 processes around \SI{310}{\mega\bit\per\second}, and another between 15 and 20 processes giving between 210 and \SI{250}{\mega\bit\per\second}.
As for the response time, unlike with S3, for scratch, it is much more predictable.
Between 2 and 20 processes, peaking 10 we have the largest response times, and considering the throughput, 2, 15 or 20 processes are the best choice since then we have the highest throughput and lowest response time.

This experiment highlights the main throughput and response time differences between using local and remote storage.
For accessing remote storage, concurrency is key.
Also, this experiment outlines what is the maximum performance we can get for image loading, using Python multiprocessing.
Considering that the throughput from the \pycode{Dataloader} experiments are much higher, for both types of storage, a combination of multiprocessing, threading and asynchronous requests provides the best data loading performance. 

%% file: chapters/ch4_all_throughputs.tex
\section{Complete end-to-end benchmark}

Each of the previous experiments addresses a certain aspect for dealing with remote storage and concurrency.
The evaluated modifications include a concurrency layer in the fetcher, additional thread pool for the workers (only in Threaded implementation), lazy initialization of the Dataloader, and cashing.
Furthermore, the maximum throughput of each data loading layer has been explored.
In this section, we present the initial experiment (see \autoref{tab:intro_motivation_example}) repeated with all the aforementioned modifications. 

\begin{figure}[ht!]
\includegraphics[width=0.65\textwidth]{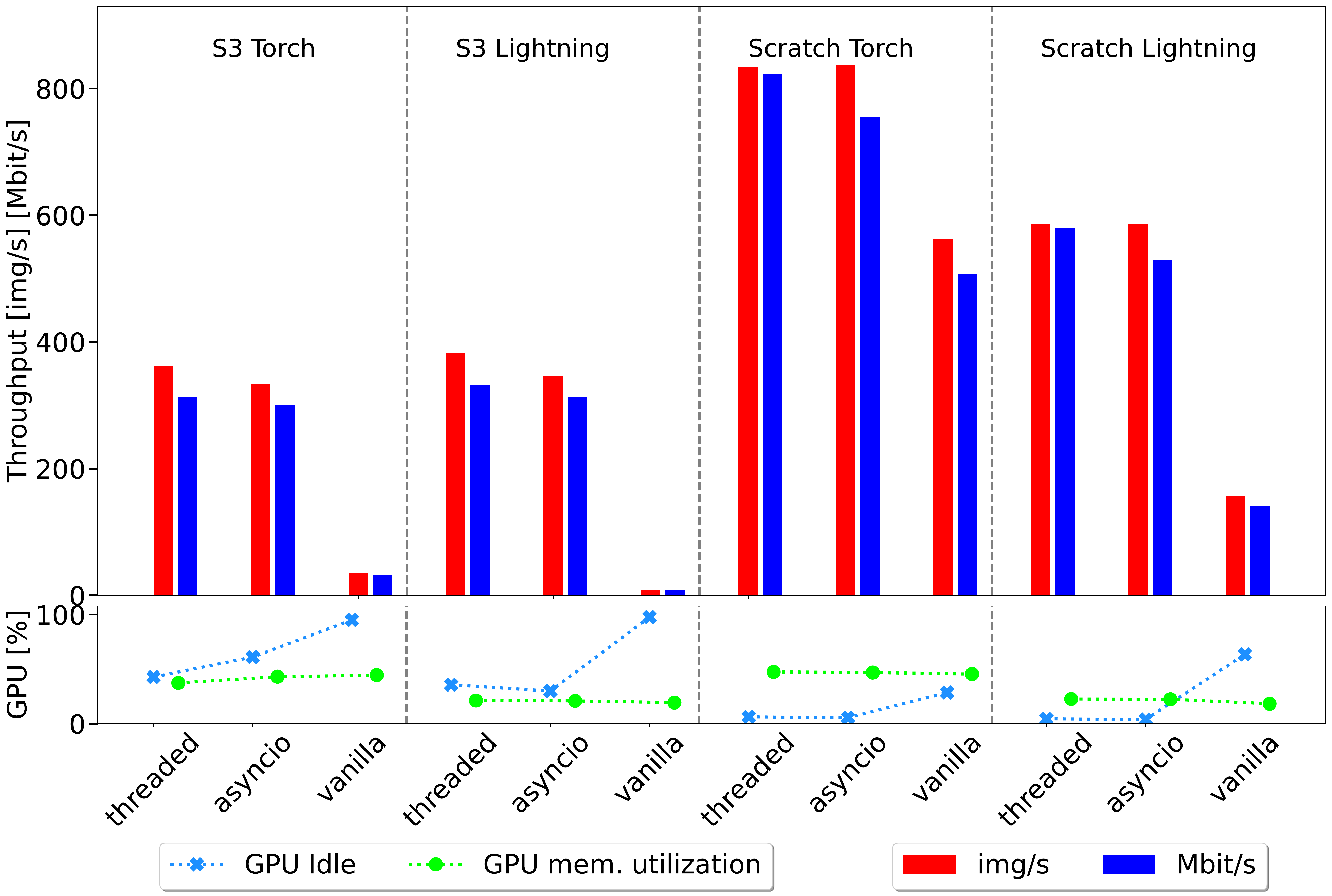}
\centering
\caption{Initial experiment repeated, with all the introduced modifications, but with original parameters to keep the experiments consistent and comparable to the motivational experiment. S3 vs Scratch, with Threaded, Asyncio and Vanilla implementation, including the GPU idling and memory utilization.}
\label{fig:ch4/final_measurements}
\end{figure}

\autoref{fig:ch4/final_measurements} shows the results, for all combinations of implementations and libraries, including the throughput and GPU utilization (processing and memory).
The motivation for this work is to approach the throughput of local storage, and increase the GPU utilization, when loading data from a remote storage.
Even though the introduced modifications do not outperform local storage, they lead to a substantial increase in terms of throughput.
With the modifications introduced in the Threaded Torch implementation to load data from S3 it is possible to reach \SI{67}{\percent} of the Vanilla Torch implementation with respect to loading data from scratch.
This is a $15.5\times$ improvement compared to the Vanilla Torch implementation.
With Lightning, it is possible to achieve similar performance.
Even more, it is possible to outperform Lightning scratch with the Lightning Threaded implementation $2.5\times$ .
In addition to these results, \autoref{fig:ch4/final_measurements} indicates that the proposed modifications also help in improving the performance when using local storage.

The GPU processing and memory utilization, as depicted in \autoref{fig:ch4/final_measurements}, shows that the GPU idle time is significantly reduced with our modifications, in particular when using remote storage.
For GPU memory utilization there are only slight differences, which is expected given that the model itself, nor batch sizes were changed.

\begin{figure}[ht!]
\includegraphics[width=0.85\textwidth]{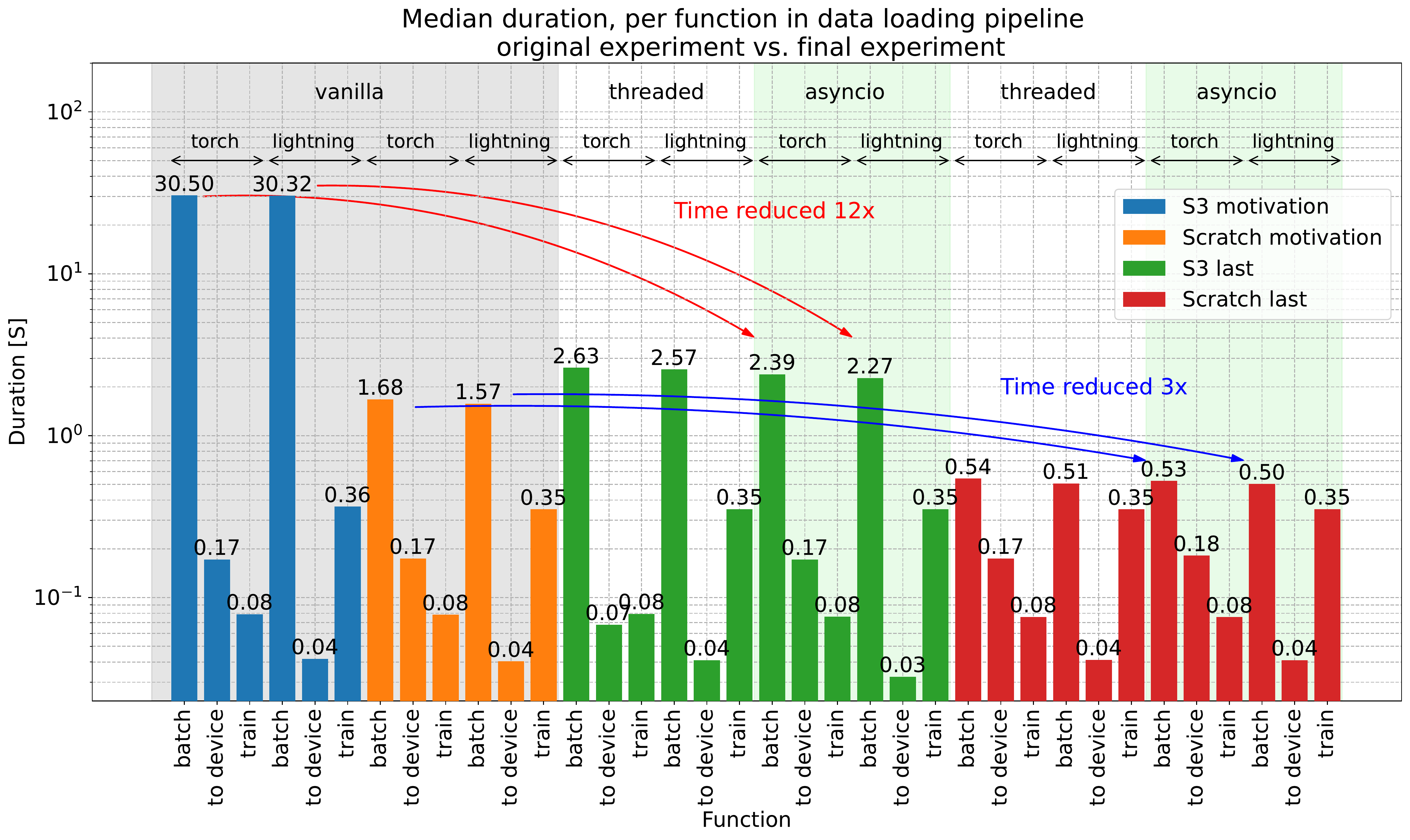}
\centering
\caption{The figure highlights significant improvements for both S3 and Scratch storage that can be achieved with modifications to the data loading pipeline introduced in this work.}
\label{fig:ch4/throughput-batches-final}
\end{figure}

\autoref{fig:ch4/throughput-batches-final} shows the median duration of the main functions used for training. The figure highlights significant improvements in batch loading.
This improvements lead to a reduction of batch loading time of up to $12\times$ for S3 cloud storage, and up to $3\times$ for Scratch storage. 

\subsection{End-to-end throughput per data loading layer}

\begin{figure}[ht!]
\includegraphics[width=0.99\textwidth]{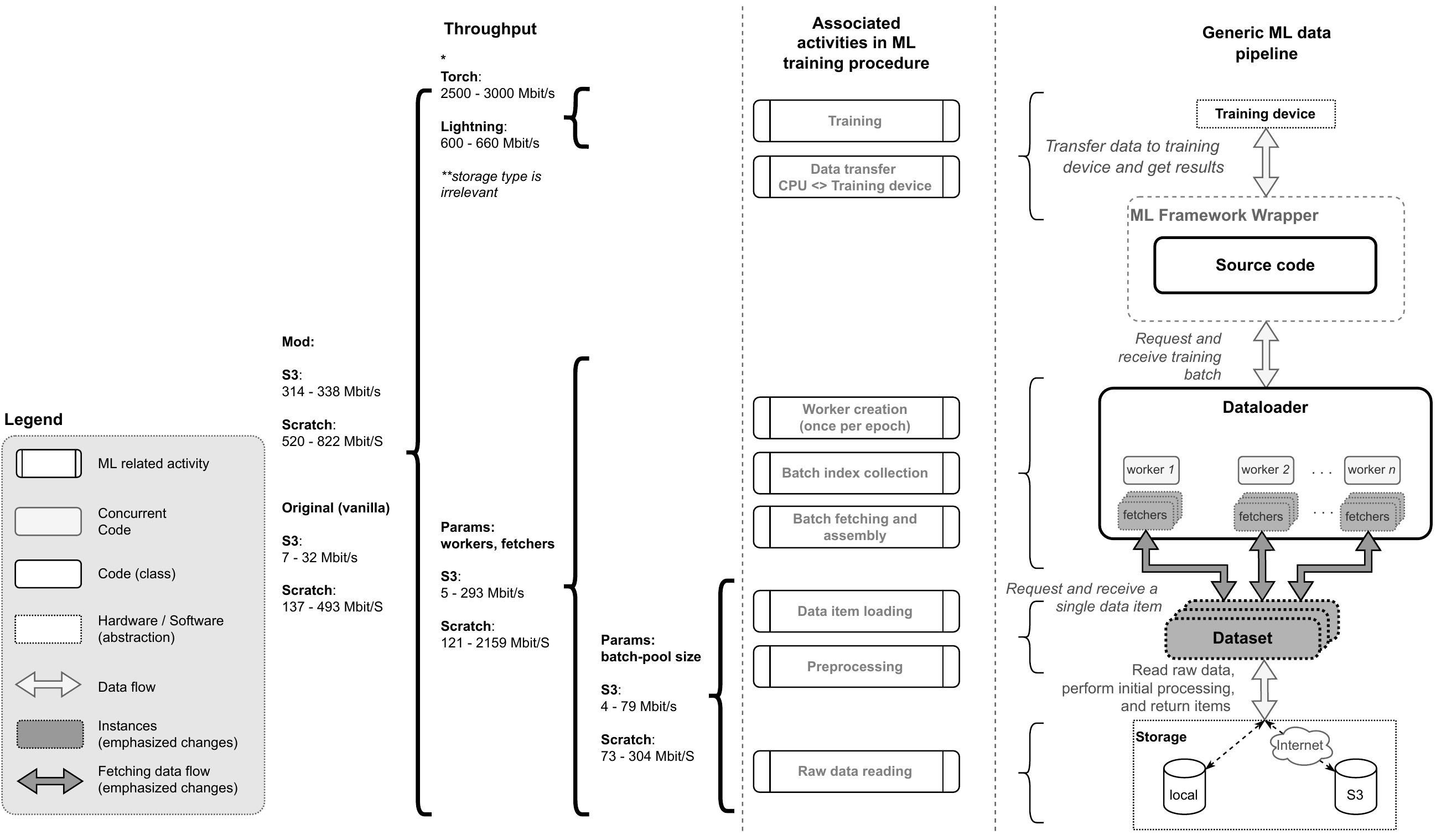}
\centering
\caption{The image highlights the changes and the throughout the data loading pipeline, along with the throughput of the training phase (* which is further addressed in the appendix) Values shown here show the throughput range which depend on different implementations (e.g. Threaded vs Asyncio, and Threaded vs Lightning).}
\label{fig:ch4/throughput-all}
\end{figure}

With the variety of presented experiments we are able to compose a figure that shows throughput rates per layer in the data loading pipeline.
\autoref{fig:ch4/throughput-all} is the extension of \autoref{fig:dataloader-stack-full} that clearly shows how it is possible to benefit from concurrency.
Starting with the pure \pycode{Dataset} with concurrency at the bottom, we can expect throughput between $4$ and \SI{79}{\mega\bit\per\second} for S3, and between $73$ and \SI{304}{\mega\bit\per\second} for scratch.
In the next layer up, with the \pycode{Dataloader}, that combines threading and multiprocessing, this significantly increases, so we can get between $5$ and \SI{293}{\mega\bit\per\second} for S3 and between $121$ and \SI{2159}{\mega\bit\per\second} for scratch.

However, for the end-to-end which includes the entire training process this gap closes down a bit, given that for S3 we get throughput between $314$ and \SI{338}{\mega\bit\per\second}, and for scratch between $520$ and \SI{822}{\mega\bit\per\second}.
Note, that for scratch, we also included numbers that use our modified data loading stack.
Compared to the original numbers, as already stated preciously, we can get up to \SI{68.5}{\percent} of scratch throughput performance.

Given the numbers presented above, one can notice that while the highest throughput for scratch storage is \SI{2159}{\mega\bit\per\second} on the \pycode{Dataloader} layer, with end-to-end the highest throughput was \SI{822}{\mega\bit\per\second}, which can indicate that the training bottleneck is no longer in the data loading pipeline. The details of the training step are addressed in depth in the appendix. 


%% file: chapters/ch5_disussion.tex
\subsection{Related work}
\label{related-work}

The related work on this subject is relatively scarce and published in a variety of venues.
This indicates that ML engineering does not quite get the attention that it deserves.
Nevertheless, more frequently journals and conferences call for special issues and tracks on ML engineering.
The relevant publications can fit into two major categories, a) ones dealing with data and case specific datasets (e.g.  Omnidata~\cite{Eftekhar_2021_ICCV} and Kaolin~\cite{jatavallabhula2019kaolin} focusing  on 3D data, TorchMeta~\cite{deleu2019torchmeta} focusing on few-shot learning, TorchIO~\cite{perez2021torchio} focusing on medical datasets,etc.), b) those emphasizing importance of efficient data loading, with the latter being the most related to the present work. 

Yang and Chong present an analytical model that helps analyze the cost of data loading in a distributed training environment~\cite{8990455}.
The authors also mention and partially exploit the untapped parallelism for loading single batches, similarly as presented in this paper.
However, the main focus of their work is aggregated caching, or \textit{locality-aware caching}, that provides a way for individual nodes to train on the data that is already stored in the local cache.
The authors report minimizing the overall data loading time and a $30\times$ speedup in data loading (with 256 nodes).
Furthermore, the authors point out that in their case Python I/O operations were chosen such that they release GIL.
Though in our work we do not address distributed training, Yang and Chong provide a roadmap for our work in that direction. 

Another important issue, with data loading, is data augmentation which is the focus of work by Zolnouri et. al.~\cite{Zolnouri2020ImportanceOD}.
The authors report that heavy data augmentation has a great impact on the standard PyTorch Dataloader, utilizing up to 40\% of the training time in their use case.
Using Nvidia's DALI\footnote{\url{https://developer.nvidia.com/dali}, a portable, open-source library for decoding and augmenting images, videos and speech to accelerate deep learning applications} they were able to improve the loading of data from Imagenet by a factor of 100.
With DALI, the data loading process is shared between CPU and GPU, and therefore the augmenting operations can be run on both.
This further confirms the issue with the PyTorch data loading pipeline. 

Furthermore, in work by Aziman et.al.~\cite{Aizman2019HighPI} from Nvidia, a great overview of data loading techniques is presented and argued how efficient data loading is essential for state-of-the-art deep learning.
The authors present a new data loading pipeline called AI Store, which seeks to address the drawbacks of distributed existing filesystems
In their work the authors point out that these are not made for data access patterns used in DL.
This includes iterating over a random permutation of a training dataset.
Also, the authors call for a framework-agnostic data loading solution to optimize for DL requirements.

In this work we have shown empirically how the Torch \pycode{Dataloader} can benefit from additional layers of parallelism.
The presented increase of the \pycode{Dataloader} efficiency due to the implemented modifications opens a gateway towards effective use of datasets stored on remote servers and object stores (e.g. AWS S3, Google Cloud, Azure Storage, etc.) to improve and simplify the dataset creation, management, and distribution. However, by using shading, FastAI and WebDataset can outperform our Dataloader (Appendix, \autoref{fastai-vs-dataset}). Whether someone wants or can use sharding is beyond this discussion, however for our future work it should be a consideration. 

%% file: chapters/ch6_conclusion.tex
\section{Conclusion and Future Work}

\paragraph{Contribution}
This technical report shows that data loading throughput can be increased by the introduction of additional concurrency and some minor changes the  PyTorch \pycode{Dataloader}, in particular in high-latency settings as when data is loaded from remote storage. 
Specifically, on a vanilla vision dataset and model, we  increased the end-to-end throughput performance by up to $15.5\times$ for fetching from S3, corresponding to 67\% of the vanilla PyTorch implementation from local SSD; batch loading time could be reduced by up to $12\times$ for S3 cloud
storage, and up to $3\times$ for Scratch storage.

\paragraph{Limitations and Future work}
Such improvement opens up the possibility to replace local, i.e. on-site storage with remote cloud-based storage, and allow for: a) creating a centralized data storage registry that wouldn't require making needless data copies, b) further decouple data and data loading from the ML framework. 

In our solution, we did not consider additional operations on the computing platform or datasets (such as, caching, sharding, data unpacking, etc.). 
With such additional techniques the end-to-end throughput could be even better. That said, we are incentivized to further explore this work, and in particular compare it to similar platforms like DALI, Ray, Tensorflow, etc. To what measure do those platforms rely on on-disk staging, dataset modifications, dataset preprocessing on the remote-storage side, etc? To what degree can our cache-free and shard-free compete with existing solutions? 

Furthermore, the implementation in a lower-level language should allow for more efficient access to networking resources, parallelization  (see appendix, \autoref{gil}) and potentially low-level communication features between the CPU and GPU. Of course, a framework like this should be accessible easily through Python, as it became a de-facto domain-specific language of machine learning.







%% file: chapters/appendix.tex
\clearpage

\section{Appendix}

\subsection{Exhaustive benchmark of different storage}
\label{apx:exhaustive-benchmark}
\begin{figure}[ht!]
\includegraphics[width=0.65\textwidth]{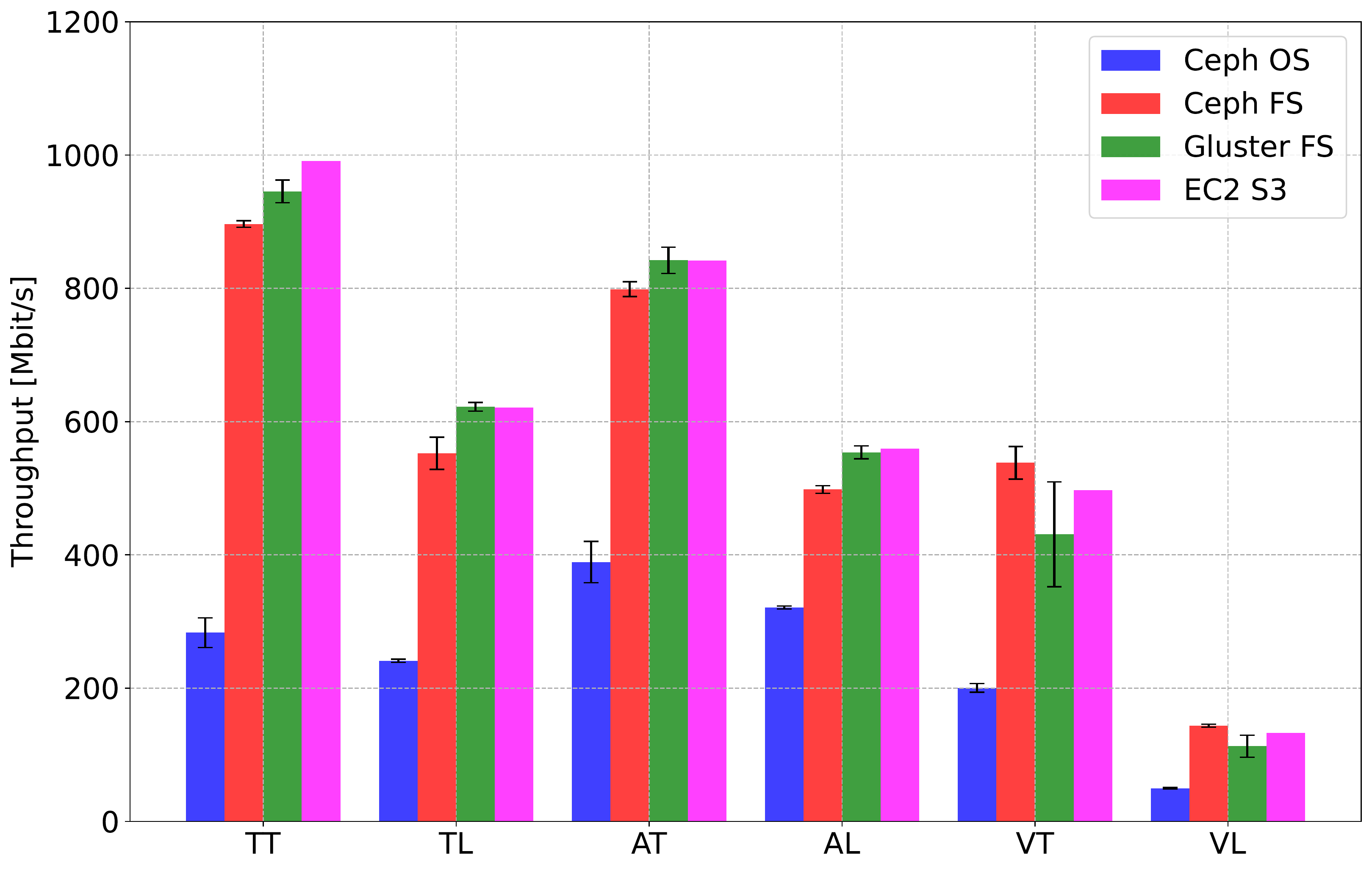}
\centering
\caption{Throughput (\si{\mega\bit\per\second}) comparison between different storage types. Gluster FS (Datacenter 1) and Ceph FS (Datacenter 2) experiments are repeated 10 times, Ceph OS 6 times (due to its duration, on Datacenter 1), and S3 with EC2 machine, only once. The $x$-axis shows abbreviations for libraries and implementations (e.g. \textit{TT} is Torch Threaded, \textit{TL} is Torch Lightning, \textit{AT} is Asyncio Torch, etc.)}
\label{fig:appendices/fs-benchmarks}
\end{figure}

\autoref{fig:appendices/fs-benchmarks} shows the average throughput, along with error bars for an exhaustive benchmark performed on different storage types. The parameters were the following:

\begin{table}[ht!]
\centering
\resizebox{0.7\textwidth}{!}{
\begin{tabular}{@{}lllllllll@{}}
\toprule
\textbf{\begin{tabular}[c]{@{}l@{}}Batch \\ size\end{tabular}} &
  \textbf{Workers} &
  \textbf{\begin{tabular}[c]{@{}l@{}}Prefetch\\ factor\end{tabular}} &
  \textbf{\begin{tabular}[c]{@{}l@{}}Number of \\ fetchers\end{tabular}} &
  \textbf{\begin{tabular}[c]{@{}l@{}}Batch\\ pool\end{tabular}} &
  \textbf{\begin{tabular}[c]{@{}l@{}}Dataset\\ limit (size)\end{tabular}} &
  \textbf{\begin{tabular}[c]{@{}l@{}}Learning\\ rate\end{tabular}} &
  \textbf{\begin{tabular}[c]{@{}l@{}}Weight\\ decay\end{tabular}} &
  \textbf{Epochs} \\ \midrule
\multicolumn{1}{c}{64} &
  \multicolumn{1}{c}{4} &
  \multicolumn{1}{c}{2} &
  \multicolumn{1}{c}{16} &
  \multicolumn{1}{c}{512} &
  \multicolumn{1}{c}{35000} &
  \multicolumn{1}{c}{0.1} &
  \multicolumn{1}{c}{0.0001} &
  \multicolumn{1}{c}{100} \\ \bottomrule
\end{tabular}
}
\caption{Parameters for exhaustive benchmarking}
\label{tab:exhaustive-benchmark-params}
\end{table}

To compensate for the fade-in and fade-out effect (see \autoref{fade-in-fade-out}), the experiments were set to run for a longer period than previous ones (100 epochs). For Ceph object store (Ceph OS) and Ceph file system (Ceph FS) we used the Datacenter 2, for the Gluster file system (Gluster FS) we used the Datacenter 1, and finally, for S3 we used AWS EC2 instance. 

To get an idea about the measurement error, the experiments with Gluster FS and Ceph FS were repeated 10 times to, while the Ceph OS was repeated only 6 times, due to its long duration. For clarity, the Vanilla Lightning experiment with Ceph OS ran for 18 hours, and repeating it 6 times meant runtime of 4.5 days. Just for a single experiment. For similar reasons, the experiment with EC2 using S3 (object store), was not repeated, and thus, doesn't show the error bars.  

For most storage types, that is, for Gluster FS, Ceph FS, and S3, the results are similar, however, for Ceph OS the throughput was significantly lower. Also, we can see that for the significantly longer benchmarks, compared to previous ones, the modifications in the data loading stack make a significant difference compared to the Vanilla version. 


\subsection{Sanity check: Google Colab}

We ran a similar experiment on the Google Colab platform. Again, we used AWS S3, however, we weren't able to run multiple experiments due to several usage limits warnings: \textit{GPUs and TPUs are sometimes prioritized for users who use Colab interactively rather than for long-running computations, or for users who have recently used less resources in Colab. As a result, users who use Colab for long-running computations, or users who have recently used more resources in Colab, are more likely to run into usage limits and have their access to GPUs and TPUs temporarily restricted.}
That said, we reduced the number of epochs, and ran a single experiment with the following parameters (for pure Torch library, with Threaded, Asyncio, and Vanilla implementation): 

\begin{table}[ht!]
\centering
\resizebox{0.7\textwidth}{!}{
\begin{tabular}{@{}lllllllll@{}}
\toprule
\textbf{\begin{tabular}[c]{@{}l@{}}Batch \\ size\end{tabular}} &
  \textbf{Workers} &
  \textbf{\begin{tabular}[c]{@{}l@{}}Prefetch\\ factor\end{tabular}} &
  \textbf{\begin{tabular}[c]{@{}l@{}}Number of \\ fetchers\end{tabular}} &
  \textbf{\begin{tabular}[c]{@{}l@{}}Batch\\ pool\end{tabular}} &
  \textbf{\begin{tabular}[c]{@{}l@{}}Dataset\\ limit (size)\end{tabular}} &
  \textbf{\begin{tabular}[c]{@{}l@{}}Learning\\ rate\end{tabular}} &
  \textbf{\begin{tabular}[c]{@{}l@{}}Weight\\ decay\end{tabular}} &
  \textbf{Epochs} \\ \midrule
\multicolumn{1}{c}{64} &
  \multicolumn{1}{c}{4} &
  \multicolumn{1}{c}{2} &
  \multicolumn{1}{c}{16} &
  \multicolumn{1}{c}{256} &
  \multicolumn{1}{c}{3000} &
  \multicolumn{1}{c}{0.1} &
  \multicolumn{1}{c}{0.0001} &
  \multicolumn{1}{c}{5} \\ \bottomrule
\end{tabular}
}
\caption{Parameters for Google Colab benchmark}
\label{tab:colab-params}
\end{table}

We were able to only measure the experiment duration, however given that the average image size is \SI{115}{\kilo\byte} we were able to (roughly) infer the throughput from this. The results are shown in the table:

\begin{table}[ht!]
\centering
\resizebox{0.85\textwidth}{!}{
\begin{tabular}{@{}lccccc@{}}
\toprule
\textbf{Implementation} &
  \textbf{\begin{tabular}[c]{@{}c@{}}Time \\ {[\si{\second}]}\end{tabular}} &
  \textbf{\begin{tabular}[c]{@{}c@{}}Total images \\ $(dataset \cdot epochs)$\end{tabular}} &
  \textbf{\begin{tabular}[c]{@{}c@{}}Throughput {[\si{\img\per\second}]}\\ $(total\mbox{ }images / time)$\end{tabular}} &
  \textbf{\begin{tabular}[c]{@{}c@{}}Handled data size {[\si{\mega\bit}]}\\ \SI{1}{\img}$=$\SI{115}{\kilo\byte}$=$\SI{0.92}{\mega\bit}\end{tabular}} &
  \textbf{\begin{tabular}[c]{@{}c@{}}Throughput\\ {[\si{\mega\bit\per\second}]}\\ $(data size / time)$\end{tabular}} \\ \midrule
\textbf{Asyncio}  & 263.08 & 15000 & 57.02 & 13800 & 52.46 \\
\textbf{Threaded} & 264.16 & 15000 & 56.78 & 13800 & 52.24 \\
\textbf{Vanilla}  & 385.93 & 15000 & 38.87 & 13800 & 35.76 \\ \bottomrule
\end{tabular}
}
\vspace{2mm}
\caption{Experiment runtime on Google Colab platform shows that with the modified data loading stack, we were able to benefit from an extra layer of concurrency. The table also shows the inferred throughput.}
\vspace{-5mm}
\label{tab:appendixcolab}
\end{table}

\subsection{Lightning vs Torch}

\subsubsection{Performance differences I: data loading}

Throughput the many benchmarks performed in this work, we noticed lower performance from Lightning, than from Torch. Since Lightning is a wrapper that uses the Torch framework, it is interesting to report where the performance difference is coming from. We produced a new experiment with the following parameters to demonstrate the performance difference:

\begin{table}[ht!]
\centering
\resizebox{0.7\textwidth}{!}{
\begin{tabular}{@{}lllllllll@{}}
\toprule
\textbf{\begin{tabular}[c]{@{}l@{}}Batch \\ size\end{tabular}} &
  \textbf{Workers} &
  \textbf{\begin{tabular}[c]{@{}l@{}}Prefetch\\ factor\end{tabular}} &
  \textbf{\begin{tabular}[c]{@{}l@{}}Number of \\ fetchers\end{tabular}} &
  \textbf{\begin{tabular}[c]{@{}l@{}}Batch\\ pool\end{tabular}} &
  \textbf{\begin{tabular}[c]{@{}l@{}}Dataset\\ limit (size)\end{tabular}} &
  \textbf{\begin{tabular}[c]{@{}l@{}}Learning\\ rate\end{tabular}} &
  \textbf{\begin{tabular}[c]{@{}l@{}}Weight\\ decay\end{tabular}} &
  \textbf{Epochs} \\ \midrule
\multicolumn{1}{c}{256} &
  \multicolumn{1}{c}{4} &
  \multicolumn{1}{c}{2} &
  \multicolumn{1}{c}{16} &
  \multicolumn{1}{c}{512} &
  \multicolumn{1}{c}{2048} &
  \multicolumn{1}{c}{0.1} &
  \multicolumn{1}{c}{0.0001} &
  \multicolumn{1}{c}{1} \\ \bottomrule
\end{tabular}
}
\caption{Parameters for comparison in performance for PyTorch and Lightning}
\label{tab:torch-lightning-performance-difference-params}
\end{table}

By adding additional logging, we were able to produce a complete picture of the execution order for each batch and epochs. It is displayed in \autoref{fig:appendices/lightning-steps} below.  

\begin{figure}[ht!]
\includegraphics[width=0.60\textwidth]{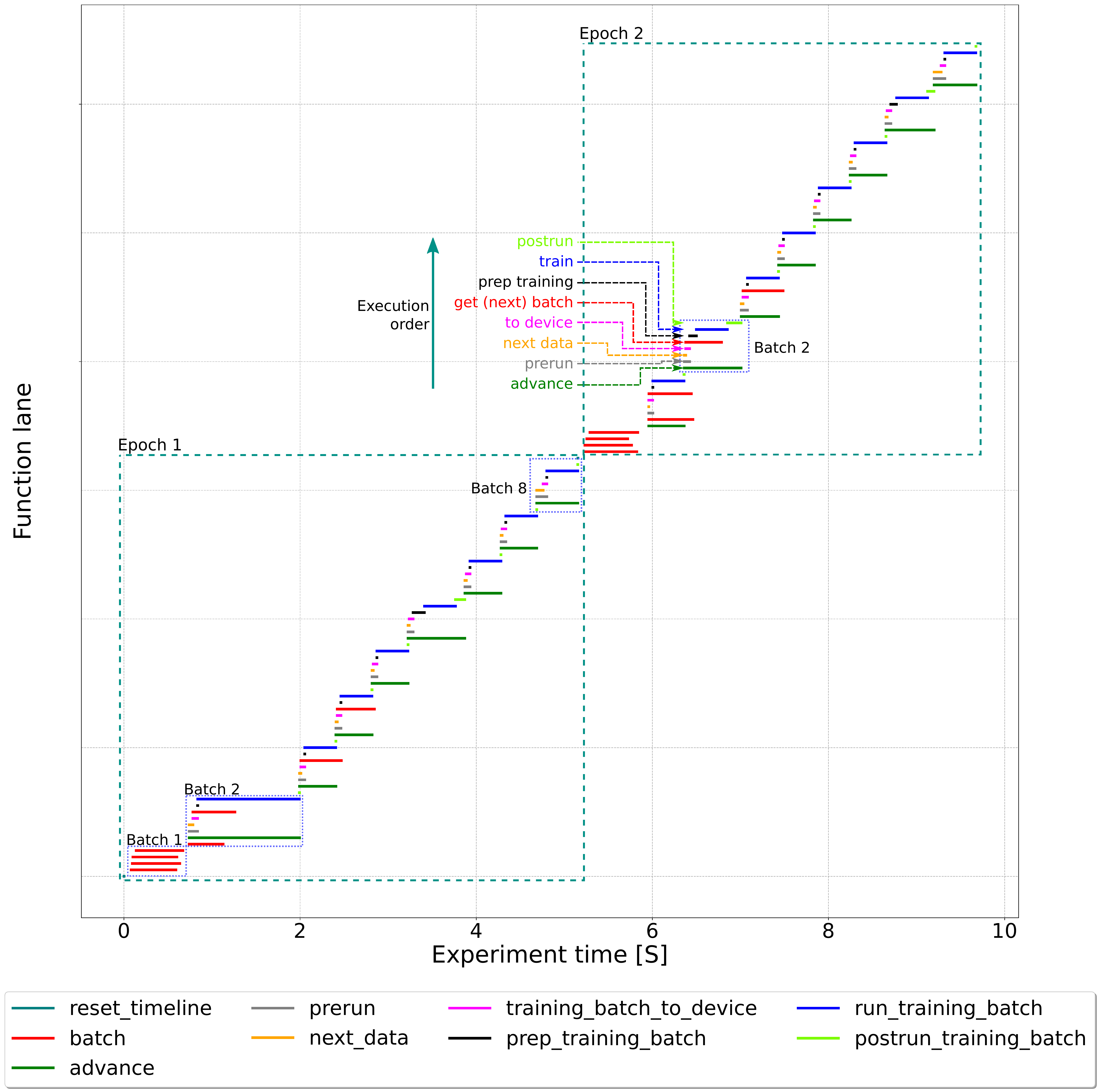}
\centering
\caption{Lightning function execution order, elaborated on a single batch.}
\label{fig:appendices/lightning-steps}
\end{figure}

The image shows Lightning implementation with two epochs, where each one progresses through 8 batches. There are 4 workers, and the prefetch factor is 2. With our modifications, as soon as workers are created, each one gets a batch to download, so, initially we see 4 read lines where each one represents a worker collecting a batch. In this case, the prefetch factor 2 means that we need at least two batches before continuing to the training process. Now, let’s consider a highlighted batch in the Epoch 2:

\begin{itemize}
    \item \textit{Advance}, is the lane that indicates the run of the Lightning function call advance which uses a single batch to train. This is a function containing all the subsequent steps. 
    \item \textit{Prerun}, is a lane that is measured from entering the advanced function till the data is loaded to the device. In this case it encapsulates the following two lanes, next data and to device.
    \item \textit{Next data}, is the lane that shows the execution of the function \pycode{next}, that triggers the loading of the next batch. 
    \item \textit{To device,} is the lane that shows the execution of the \pycode{batch\_to\_device} function which copies the batch to the GPU memory.
    \item \textit{Get next batch}, it is a lane that shows the execution of functions involved in obtaining the training batch. It’s rather complicated under the hood, so the \autoref{fig:appendices/exec-order-lighting} illustrates the procedure. Once the next function is triggered in the advance call, it triggers the \pycode{\_\_next\_\_} function (1) in the \pycode{Dataloader}. It then starts our function to \pycode{start\_download} (2), that starts creating processes and triggered data fetching. However if it’s already downloading, it just returns (3). Then (4) the \pycode{\_next\_data}  waits until the requested data arrives. When that happens function \pycode{\_process\_data} (5) is called. It triggers the download of the next batch, and returns the data of the current one (6). That way, with continuous calls to next, we are keeping the workers busy downloading data. 
    
    \begin{figure}[ht!]
        \includegraphics[width=0.45\textwidth]{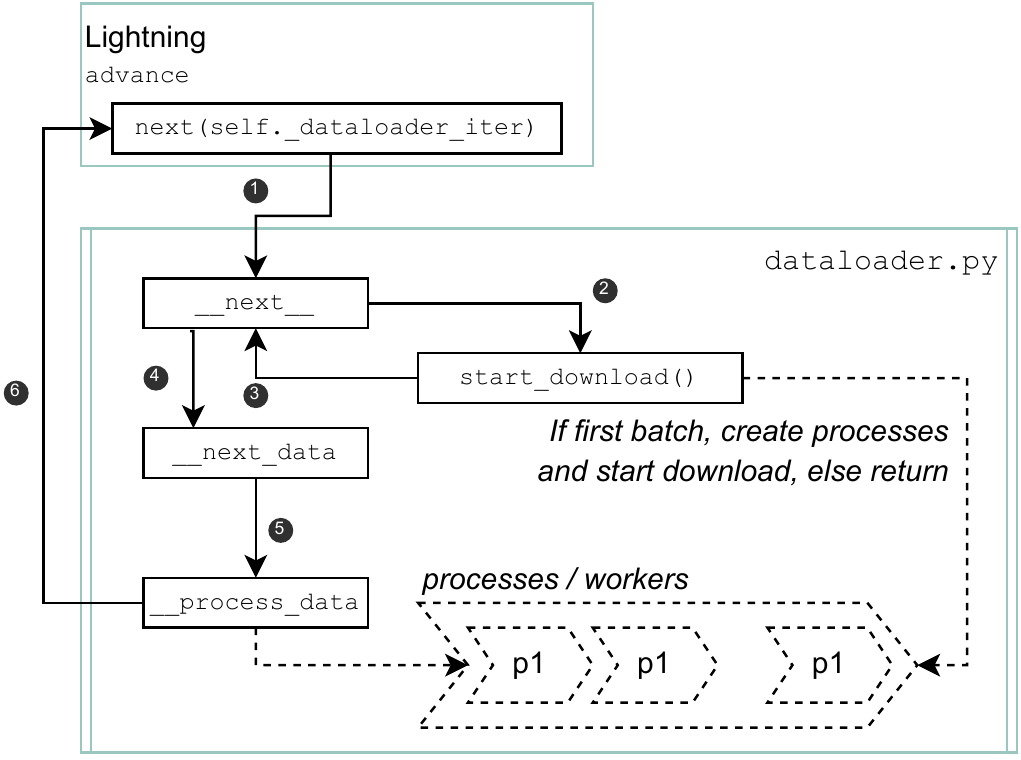}
        \centering
        \caption{Function calls and the execution order dealing with obtaining the training batch}
        \label{fig:appendices/exec-order-lighting}
    \end{figure}

    \item \textit{Prep training}, is the lane which shows the runtime of the several functions before the actual training steps, mostly involving hooks and callbacks.
    \item \textit{Train}, lane showing the training progress
    \item \textit{Postrun}, lane showing the set of functions, similar to the prep training.
\end{itemize}

\begin{figure}[hb!]
\includegraphics[width=0.60\textwidth]{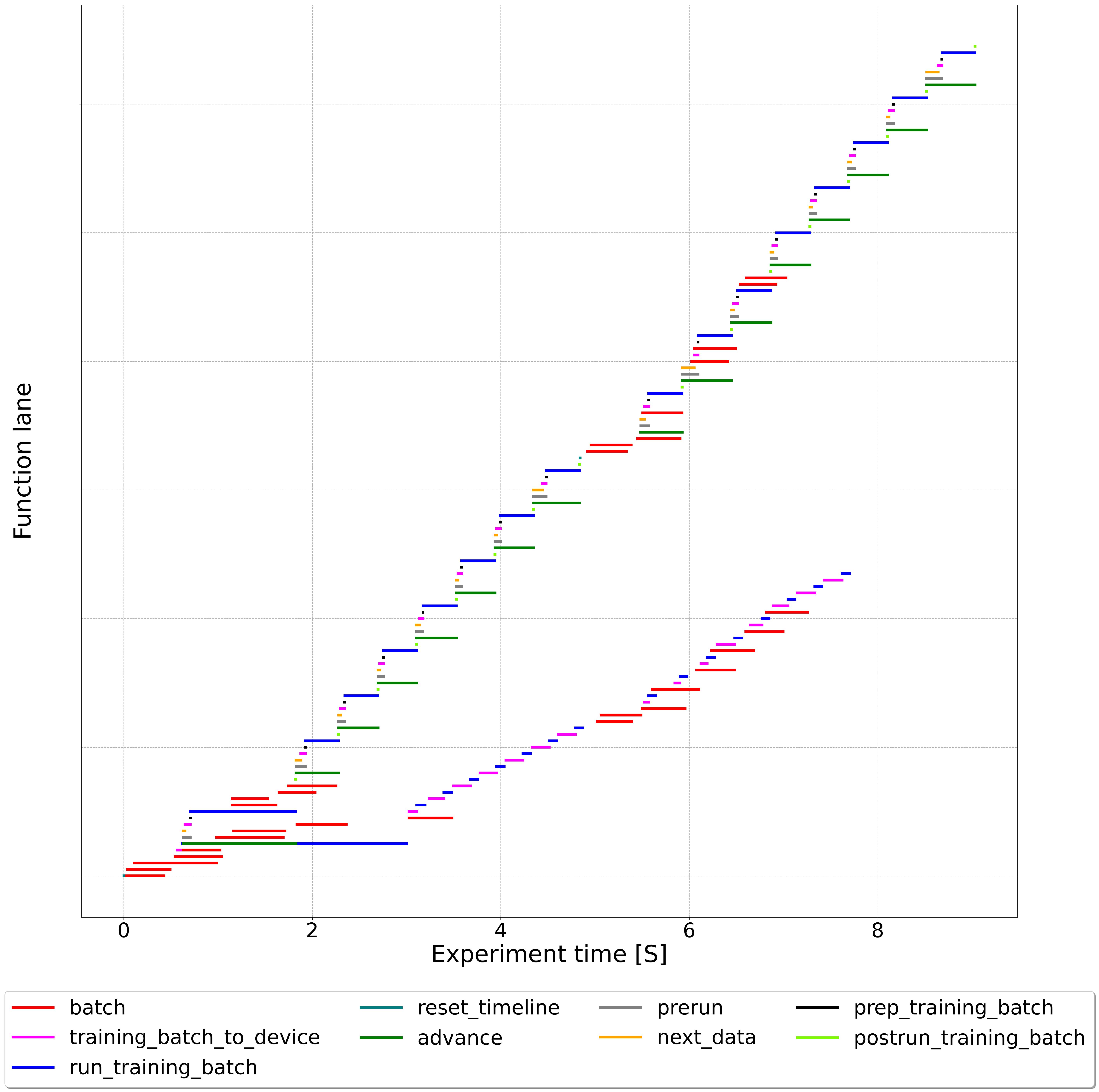}
\centering
\caption{Lightning and Torch function calls overlap.}
\label{fig:appendices/lightning-torch-overlap}
\end{figure}

Running multiple test, we noticed that sometimes the prep-training and postrun can take a significant amount of time. Tracing the code, we noticed that the following call in the advance prep-training takes the most:

\begin{lstlisting}[
    basicstyle=\ttfamily\scriptsize,
    backgroundcolor = \color{lightgray}
]
response = self.trainer.call_hook("on_train_batch_start", batch, batch_idx, **extra_kwargs)
\end{lstlisting}

\vspace{-3mm}
The longest running code in the \pycode{call\_hook} function is:

\begin{lstlisting}[
    basicstyle=\ttfamily\scriptsize,
    numbers=left,
    backgroundcolor = \color{lightgray}
]
callback_fx = getattr(self, hook_name, None)
if callable(callback_fx):
    callback_fx(*args, **kwargs) # <--- slow when "on_train_batch_start"
\end{lstlisting}




\vspace{-3mm}
With further tracing, one can get to the \pycode{callbacks/base.py}, which is implemented in \pycode{callbacks/gpu\_stats\_monitor.py}. It does nothing but triggers the logger. This indicates that we likely used slightly too aggressive approach to logging. Afterwards, we reduced the logging frequency and (\pycode{log\_every\_n\_steps}) removed the \pycode{Profiler} form our \pycode{Trainer}. This improved the Lightning performance, however, still, compared to raw Torch, in this particular case it was slightly slower. \autoref{fig:appendices/lightning-torch-overlap} shows the overlap of function timeline calls between Lightning and Torch. It's important to mention there, that the aforementioned logging features (particularly related to GPU) aren't enabled by default, meaning that it should normally not cause any performance issues. We would advise caution while chaining the default settings of Lightning profiler and its logging features.  

Pre-- and post-- training calls, though significantly shorter now, in time build up, making Lightning perform slightly worse than Torch, in this particular example. However this is not a general statement. 


\subsubsection{Performance differences II: training}

\begin{figure}[ht!]
	\begin{minipage}{0.49\linewidth}
        \centering
        \includegraphics[width=\textwidth]{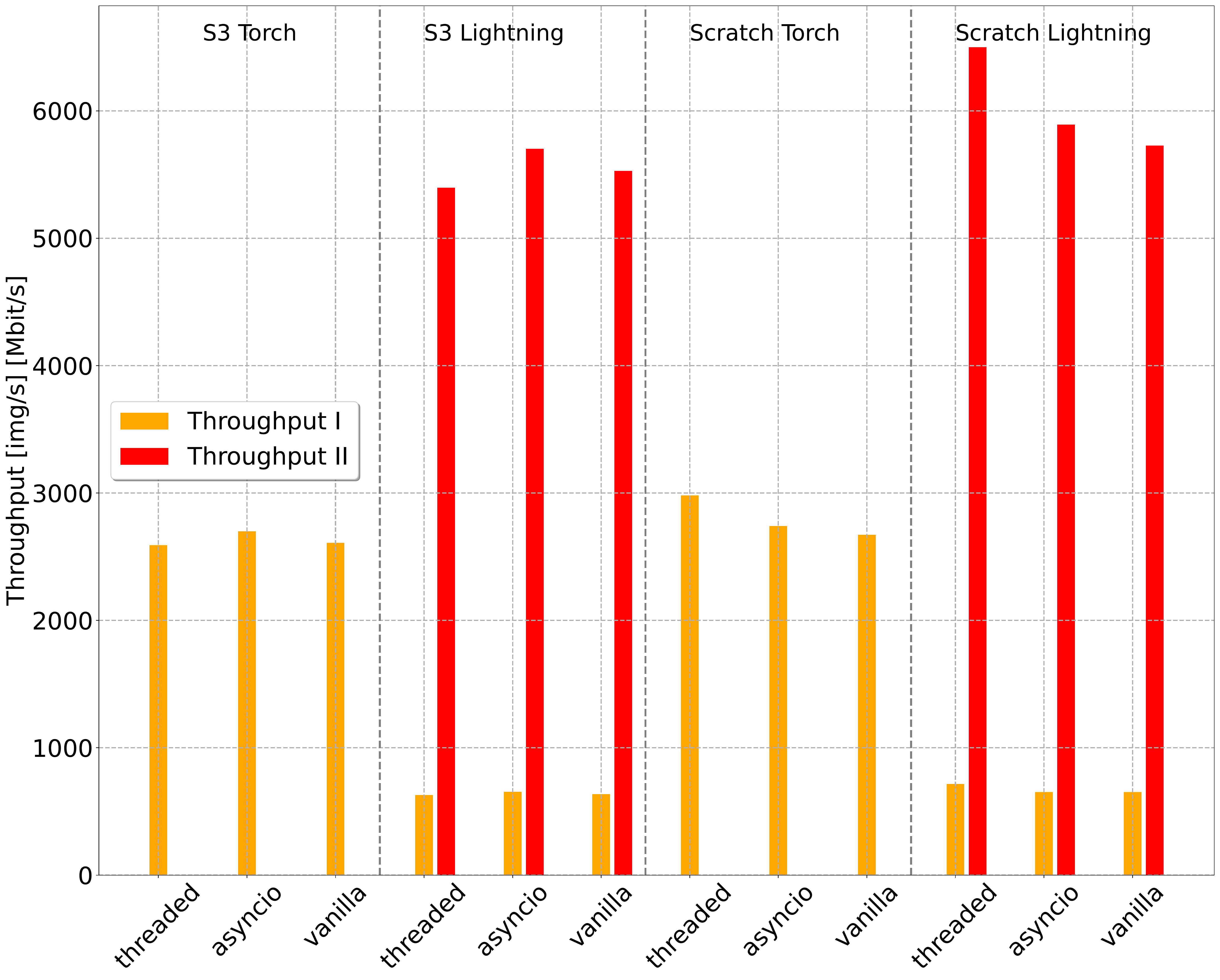} 
	\end{minipage}\hfill
	\begin{minipage}{0.49\linewidth}
	    \centering
        \includegraphics[width=\textwidth]{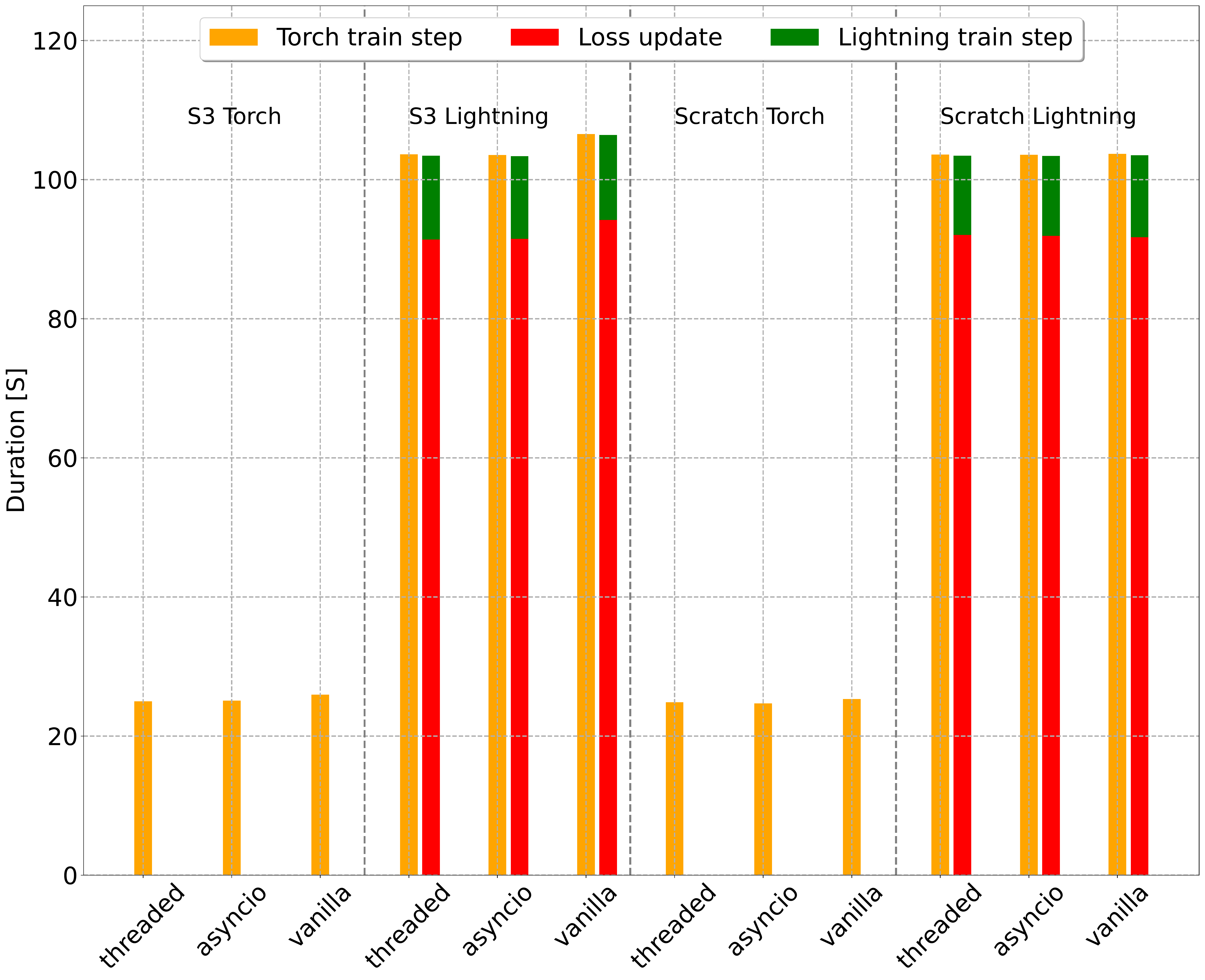} 
	\end{minipage}
    \vspace{2mm}
    \caption{Training phase throughput, for different libraries and implementations (left), along with the cumulative training duration broken down into two longest functions.}
    \label{fig:appendices/gpu-throughput}
\end{figure}

\autoref{fig:appendices/gpu-throughput} shows the throughput and the duration of the training step for Torch and Lightning, with different storage and fetch implementations. The left image shows the throughput in \si{\mega\bit\per\second} calculated as $T_{mbits} = \left(DL / \sum_{n=1}^{N} i_n \right)$ where $DL$ is the size of all the loaded data items (in \si{\mega\bit\per\second}) and $i_1, i_2, ..., i_N$ are the log entries related to the duration of \pycode{run\_training\_batch} function. That said, there are several key points to clarify here:

\begin{itemize}
    \item \textit{Throughput I}, as expected, at this stage, the data is loaded in memory, and therefore different data loading implementations and storage types should not matter. As seen by the bar plot (orange lines, \textit{Throughput I}), that is the case. 
    \item \textit{Throughput II}, instead of using the function \pycode{run\_training\_batch}, uses \pycode{optimizer\_step} which runs the training step, and the optimizer step\footnote{\url{https://github.com/PyTorchLightning/pytorch-lightning/blob/1.5.9/pytorch_lightning/loops/optimization/optimizer_loop.py\#L246-L269}}.   
    \item \textit{Torch vs Lightning}, difference comes from the fact that with Torch implementation, the measurement of GPU throughput is straightforward, as we have a clear and manual access to the forward, backward and optimization step\footnote{\url{https://github.com/pytorch/examples/blob/main/imagenet/main.py\#L289-L307}}, while with Lightning, this is hidden away by the framework generalizations (through various loops, i.e. epoch, batch, optimization and training loop). In lightning, the advance function of the \pycode{training\_batch\_loop.py}, triggers the advance step of the \pycode{training\_batch\_loop.py}, which after some additional function passing ends up in the \pycode{run\_optimization} function of the \pycode{optimizer\_loop.py}. The right side of \autoref{fig:appendices/gpu-throughput} shows the duration of the aforementioned functions, and for the Lightning the orange training step, is broken down into the training step and the loss update, which takes the majority of the time. This function comes from the automatic optimization, and with our additional tests for using the manual optimization, we got similar numbers. This means that there are also some differences in the implementation of the training step, and that the throughput can only be considered as a wide range between $650$ and \SI{3000}{\mega\bit\per\second}.
\end{itemize}

Considering the statements above, we can conclude that certain bottlenecks also exist between the data loading process, and the training step itself, which is of scope of this analysis. 

\subsection{The dreaded GIL}
\label{gil}

\begin{figure}[ht!]
\includegraphics[width=0.55\textwidth]{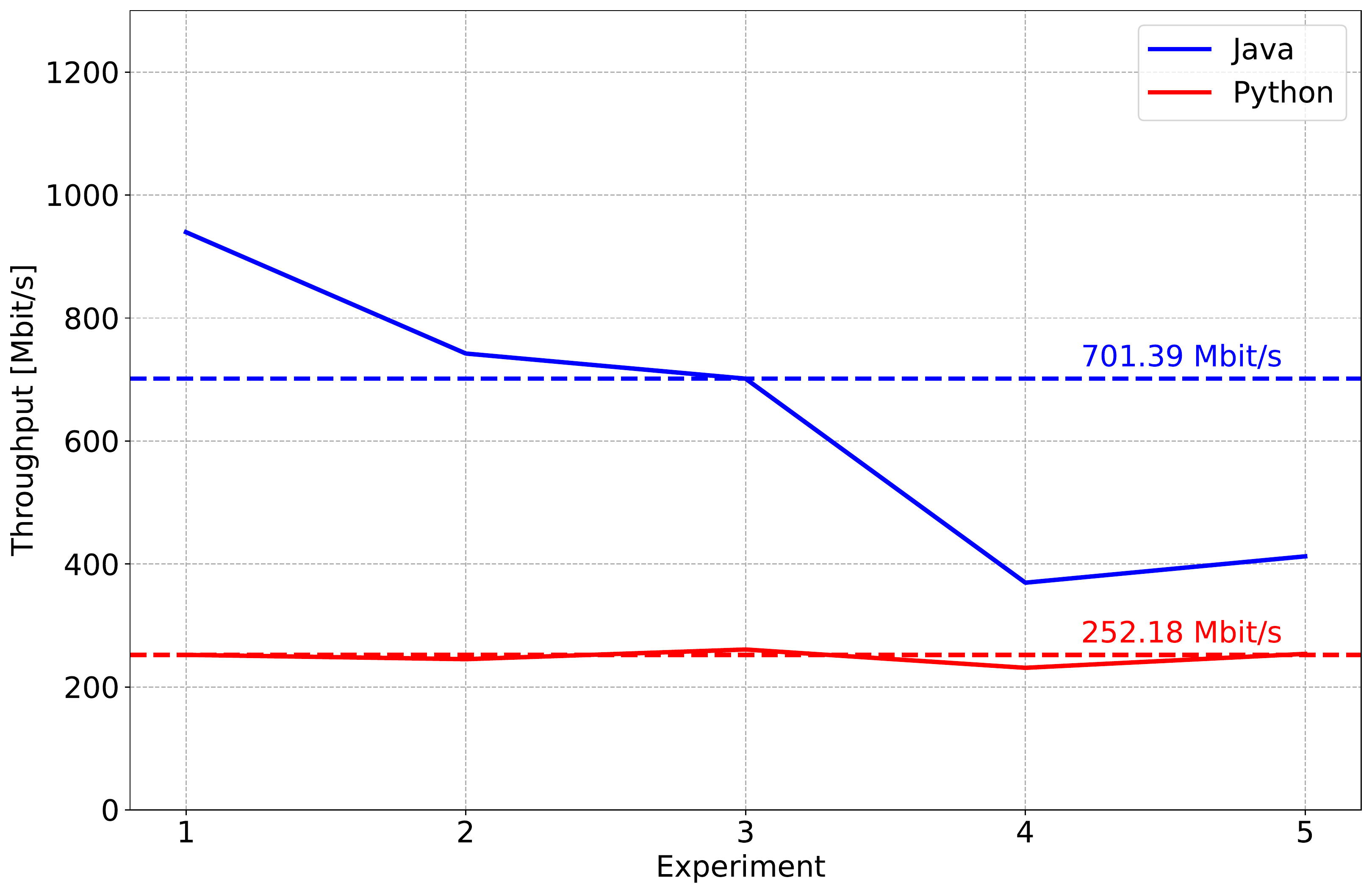}
\centering
\caption{Throughput achieved with S3 for Java and Python, both utilizing concurrency. Dashed lines show the median throughput over 5 experiments.}
\label{fig:ch4/py-vs-java}
\end{figure}

Given that we use AWS S3 as remote object store, we are using the aforementioned Boto3 Python library, and for a sanity check, we performed an experiment, where we use the pure Python multiprocessing and threading, in order to find out whether we are in fact, getting the maximum performance, and to rule out any other factors, not addressed in this work, we might have missed. In addition to this, we made the same experiment using Java. \autoref{fig:ch4/py-vs-java} shows the throughput results, for five consecutive experiments, downloading 5000 random images from S3. 

The median throughput of \SI{252.18}{\mega\bit\per\second} achieved with Python implementation roughly corresponds to the \pycode{Dataloader} performance (shown in \autoref{fig:ch4/throughput-all}, as it employs a similar combination of multiprocessing and threading. However, Java implementation achieves \SI{701.39}{\mega\bit\per\second} median throughput, which is a significant difference. Also, given the scratch performance shown in \autoref{fig:ch4/final_measurements}, with Java threading we would be able to completely replace local storage, with S3. 

That said, this indicates that we are \textit{hitting} Python's concurrency limitations, i.e. the global interpreter lock\footnote{\url{https://wiki.python.org/moin/GlobalInterpreterLock}} (GIL) prevents \textit{true} parallelism. This, in fact, is what GIL is supposed to do, it ensures that multiple threads are not executing Python bytecode at once and prevents race conditions, thus allowing for thread safety. However, it is not ideal in the sense of taking full advantage of multiprocessing systems. Though there are many pros and cons for GIL, and to Python as the language itself, we will not get into further details here, however, this is an area for why Python was not designed, and we can conclude that the concurrency layer of the data loading should be a part of an external library (e.g. Ray\footnote{\url{https://www.ray.io/}}), written in the lower-level language, like C++, similarly as is the case for lower-level layers of Torch that communicate with the GPU.

\subsection{Our implementation vs FastAI vs WebDataset}
\label{fastai-vs-dataset}

At the moment, there are several ongoing efforts that work on the aforementioned challenge.
\pycode{WebDataset}\footnote{\url{https://github.com/webdataset/webdataset}} is an example of a PyTorch \pycode{Dataset} implementation that provides access to data stored in POSIX tar archives.
It implements a standard PyTorch \pycode{IterableDataset} interface that can use local or remote tar archives, unpack them on the fly, and stream the data items into the \pycode{Dataloader}.
The \pycode{WebDataset} uses a concept called \textit{data shards}, which represent tar files containing a number (or a given size) of data items.
During training, the \pycode{WebDataset} streams unpacked data items from shards.
In the case of using remote storage, it doesn't need to keep the shards locally, as it can sequentially stream its content, with or without cache.

A very similar approach but without streaming is implemented in FastAI~\cite{info11020108}, which is a DL library developed by the fast.ai research group.
In FastAI, data is accessed via \pycode{DataBlock}, which is a class that simplifies data management in terms of data preparation and data transforms.
This \pycode{DataBlock} loads data from a given input path and is afterwards fed into the FastAI data loader.
Similar to the data loader implemented in Torch, the number of workers and batch sizes may be given as input.
The remote storage can be used through the \pycode{untar\_data} function, which downloads the entire tar file to a local path.
This path may then be used as input to the \pycode{DataBlock}.

\begin{figure}[h!]
\includegraphics[width=0.55\textwidth]{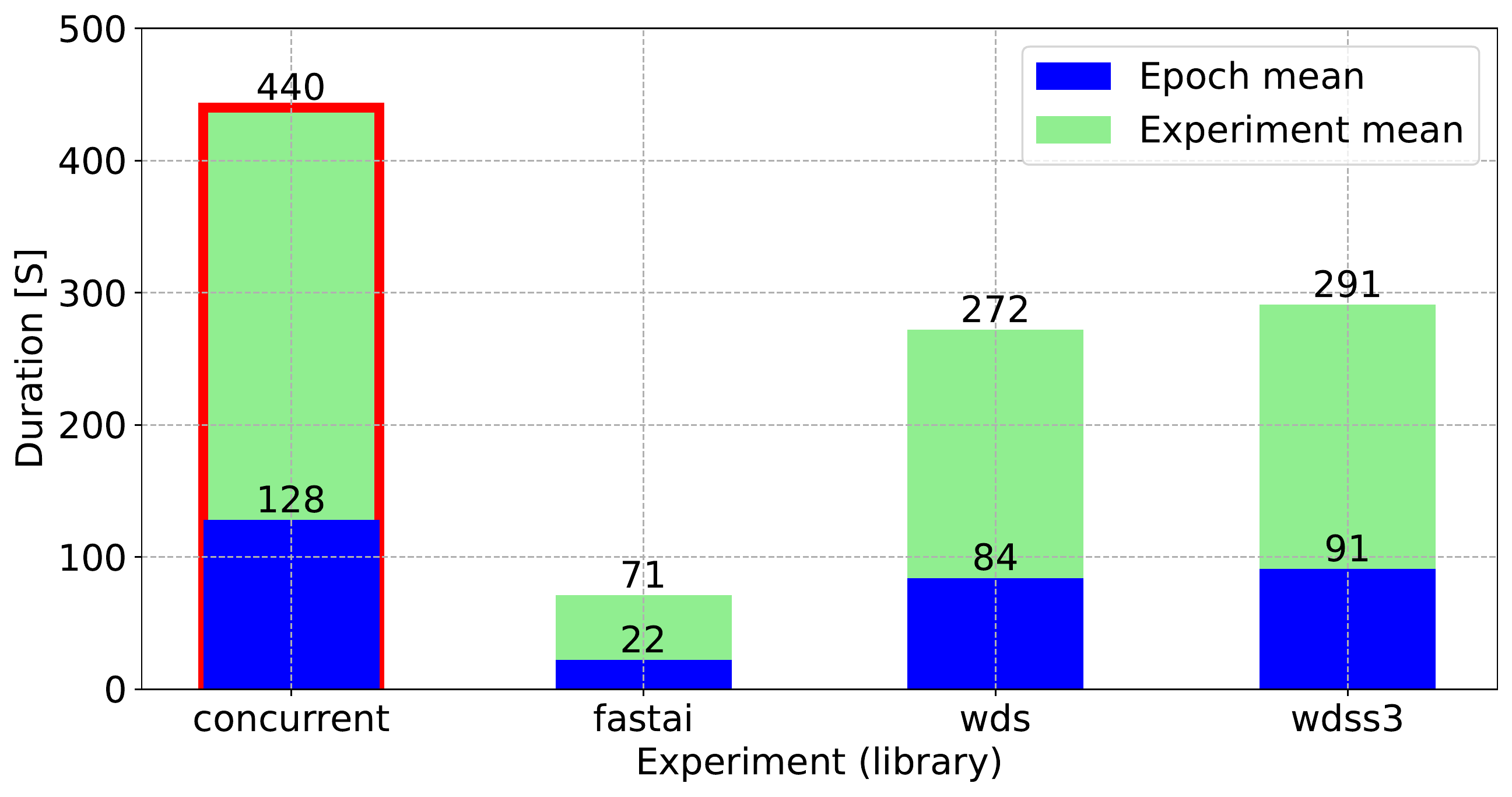}
\centering
\caption{Comparison between our implementation (concurrent), with FastAI and WebDataset. Our implementation is highlighted with the red border, and the WebDataset was used with local (\textit{wds}) and a shard stored on AWS S3 (\textit{wdss3}). The image shows the total average runtime of 5 consecutive runs, for the entire experiment and for a single epoch.}
\label{fig:ch4.5/webdataset_comparison}
\end{figure}

\autoref{fig:ch4.5/webdataset_comparison} shows the execution time comparison between the Asyncio dataloader presented in this work (\textit{concurrent}), the FastAI dataloader and WebDataset.
In case of WebDataset and FastAI, we use a single shard of 8696 images (\SI{986}{\mega\byte}).

The figure shows the average runtime for 5 consecutive runs.
It can be seen that our \pycode{ConcurrentDataset} has the highest execution time, while the FastAI implementation has the lowest.
This substantial differences in runtime may be attributed to the way these data loaders handle the data on the remote storage:
\begin{itemize}
    \item \textbf{Concurrent} For each data item a connection to S3 is established and downloaded before it is used.
    \item \textbf{FastAI} The complete tar archive is downloaded and unpacked and afterwards used by the model.
    \item \textbf{WebDataset} The data are streamed into the model and unpacked on-the-fly.
\end{itemize}
This indicates that even though our implementation delivers large improvements there are still drawbacks that need to be addressed.
This provides a guideline for future work, in which we also need to address networking overheads.

\subsection{Fade in and fade out effect}
\label{fade-in-fade-out}

\begin{figure}[ht!]
\includegraphics[width=0.95\textwidth]{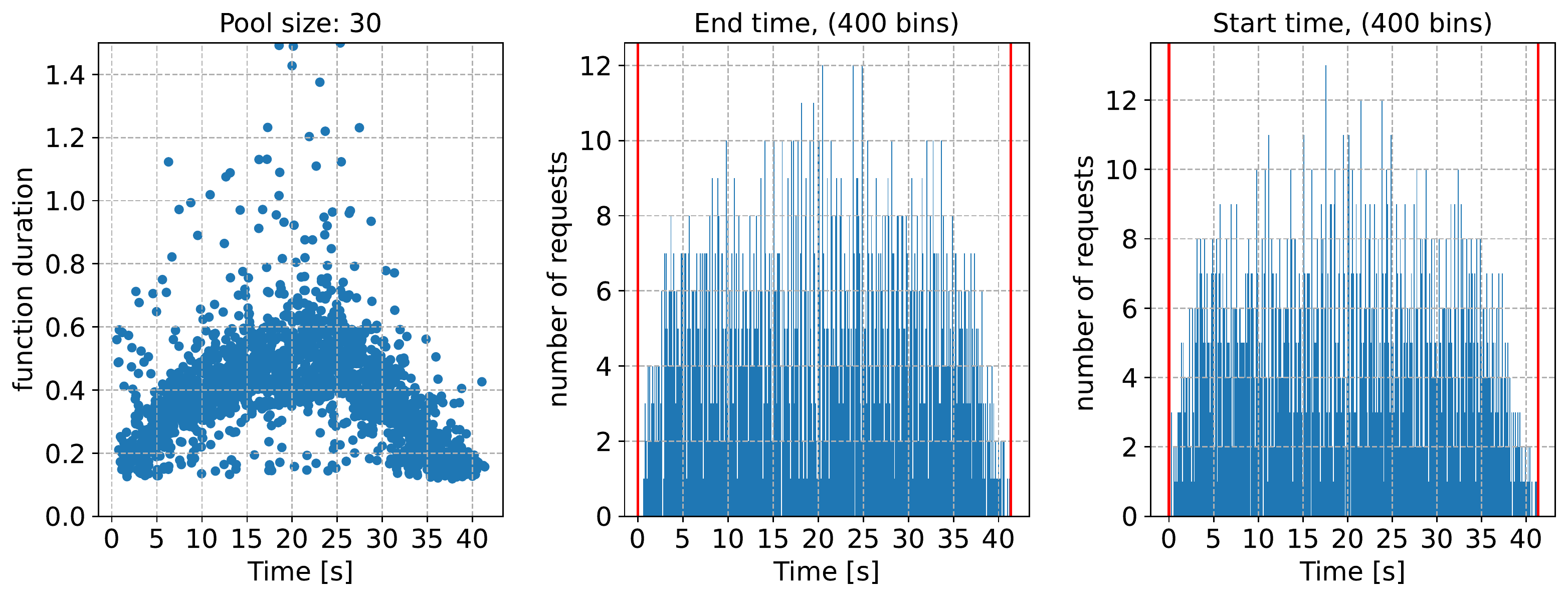}
\centering
\caption{Scatter plot of the \pycode{\_\_getitem\_\_} function start time, and its duration throughout a selected experiment (left). Histogram plot showing how many functions \pycode{\_\_getitem\_\_} finished (middle) and were started (right) at the certain point of experiment, grouped in 400 bins. The selected experiment uses S3 storage, and is \SI{40.78}{\second} long.}
\label{fig:ch3/dataset_histogram}
\end{figure}

Later on in this work, we perform additional benchmark for different types of storage.
Before proceeding, there is an important consideration that needs to be addressed.
From the previous experiment, we have chosen a specific example, that best highlights when the image loading function is called, when it finishes and how long it lasts, depending on at which point of the experiment it is called.
\autoref{fig:ch3/dataset_histogram} shows that in the beginning of the experiment we do not have many requests for image loading (fade-in), but as the experiment proceeds the requests slowly build up (fade-out, right histogram)), and similarly, at the beginning of the experiment not may images are still ready (middle histogram).
The left most scatter plot shows that at the beginning of the experiment the responses are fast, i.e. the function loading data finishes quickly, however as the experiment proceeds, it reaches a certain peak, after which it, again begins to fall down.
So, at the beginning with many requests, the responses are fast, until we saturate the multiprocessing pool, or exhaust the data source. 

This experiment demonstrates that in order for the experiments to be more convincing, they need to have a long duration, in order for fade-in and fade-out effects to become negligible.

%% file: main.bbl
\begin{thebibliography}{10}

\bibitem{8990455}
Chih-Chieh Yang and Guojing Cong.
\newblock Accelerating data loading in deep neural network training.
\newblock In {\em 2019 IEEE 26th International Conference on High Performance
  Computing, Data, and Analytics (HiPC)}, pages 235--245, 2019.

\bibitem{Aizman2019HighPI}
Alex Aizman, Gavin Maltby, and Thomas Breuel.
\newblock High performance i/o for large scale deep learning.
\newblock {\em 2019 IEEE International Conference on Big Data (Big Data)},
  pages 5965--5967, 2019.

\bibitem{NEURIPS2019_9015}
Adam Paszke, Sam Gross, Francisco Massa, Adam Lerer, James Bradbury, Gregory
  Chanan, Trevor Killeen, Zeming Lin, Natalia Gimelshein, Luca Antiga, Alban
  Desmaison, Andreas Kopf, Edward Yang, Zachary DeVito, Martin Raison, Alykhan
  Tejani, Sasank Chilamkurthy, Benoit Steiner, Lu~Fang, Junjie Bai, and Soumith
  Chintala.
\newblock Pytorch: An imperative style, high-performance deep learning library.
\newblock In {\em Advances in Neural Information Processing Systems 32}, pages
  8024--8035. Curran Associates, Inc., 2019.

\bibitem{Falcon_PyTorch_Lightning_2019}
William Falcon and {The PyTorch Lightning team}.
\newblock {PyTorch Lightning}, 3 2019.

\bibitem{he2015deep}
Kaiming He, Xiangyu Zhang, Shaoqing Ren, and Jian Sun.
\newblock Deep residual learning for image recognition, 2015.

\bibitem{deng2009imagenet}
Jia Deng, Wei Dong, Richard Socher, Li-Jia Li, Kai Li, and Li~Fei-Fei.
\newblock Imagenet: A large-scale hierarchical image database.
\newblock In {\em 2009 IEEE conference on computer vision and pattern
  recognition}, pages 248--255. Ieee, 2009.

\bibitem{hattingh2020using}
C.~Hattingh.
\newblock {\em Using Asyncio in Python: Understanding Python's Asynchronous
  Programming Features}.
\newblock O'Reilly Media, Incorporated, 2020.

\bibitem{Eftekhar_2021_ICCV}
Ainaz Eftekhar, Alexander Sax, Jitendra Malik, and Amir Zamir.
\newblock Omnidata: A scalable pipeline for making multi-task mid-level vision
  datasets from 3d scans.
\newblock In {\em Proceedings of the IEEE/CVF International Conference on
  Computer Vision (ICCV)}, pages 10786--10796, October 2021.

\bibitem{jatavallabhula2019kaolin}
Krishna~Murthy Jatavallabhula, Edward Smith, Jean-Francois Lafleche,
  Clement~Fuji Tsang, Artem Rozantsev, Wenzheng Chen, Tommy Xiang, Rev
  Lebaredian, and Sanja Fidler.
\newblock Kaolin: A pytorch library for accelerating 3d deep learning research.
\newblock {\em arXiv preprint arXiv:1911.05063}, 2019.

\bibitem{deleu2019torchmeta}
Tristan Deleu, Tobias W{\"u}rfl, Mandana Samiei, Joseph~Paul Cohen, and Yoshua
  Bengio.
\newblock Torchmeta: A meta-learning library for pytorch.
\newblock {\em arXiv preprint arXiv:1909.06576}, 2019.

\bibitem{perez2021torchio}
Fernando P{\'e}rez-Garc{\'\i}a, Rachel Sparks, and Sebastien Ourselin.
\newblock Torchio: a python library for efficient loading, preprocessing,
  augmentation and patch-based sampling of medical images in deep learning.
\newblock {\em Computer Methods and Programs in Biomedicine}, 208:106236, 2021.

\bibitem{Zolnouri2020ImportanceOD}
Mahdi Zolnouri, Xinlin Li, and V.~Nia.
\newblock Importance of data loading pipeline in training deep neural networks.
\newblock {\em ArXiv}, abs/2005.02130, 2020.

\bibitem{info11020108}
Jeremy Howard and Sylvain Gugger.
\newblock Fastai: A layered api for deep learning.
\newblock {\em Information}, 11(2), 2020.

\end{thebibliography}
